\documentclass[aip,cha,reprint]{revtex4-1}

\usepackage{amsmath}
\usepackage{amssymb}
\usepackage{graphicx}
\usepackage{dcolumn}
\usepackage{bm}
\usepackage[utf8]{inputenc}
\usepackage[T1]{fontenc}
\usepackage{mathptmx}
\usepackage{etoolbox}

\usepackage{color}

\makeatletter
\def\@email#1#2{%
 \endgroup
 \patchcmd{\titleblock@produce}
  {\frontmatter@RRAPformat}
  {\frontmatter@RRAPformat{\produce@RRAP{*#1\href{mailto:#2}{#2}}}\frontmatter@RRAPformat}
  {}{}
}%
\makeatother

\begin{document}


\title[Multifunctional physical reservoir computing in soft tensegrity robots]{
    Multifunctional physical reservoir computing in soft tensegrity robots
}

\author{Ryo Terajima}
\email[]{terajima@isi.imi.i.u-tokyo.ac.jp}
\affiliation{School of Information Science and Technology, The University of Tokyo, Tokyo, Japan.}

\author{Katsuma Inoue}
\affiliation{School of Information Science and Technology, The University of Tokyo, Tokyo, Japan.}

\author{Kohei Nakajima}
\affiliation{School of Information Science and Technology, The University of Tokyo, Tokyo, Japan.}
\affiliation{Next Generation Artificial Intelligence Research Center (AI Center), The University of Tokyo, Tokyo, Japan}

\author{Yasuo Kuniyoshi}
\affiliation{School of Information Science and Technology, The University of Tokyo, Tokyo, Japan.}
\affiliation{Next Generation Artificial Intelligence Research Center (AI Center), The University of Tokyo, Tokyo, Japan}

\date{\today}

\begin{abstract}
    Recent studies have demonstrated that the dynamics of physical systems can be utilized for the desired information processing under the framework of physical reservoir computing (PRC). Robots with soft bodies are examples of such physical systems, and their nonlinear body-environment dynamics can be used to compute and generate the motor signals necessary for the control of their own behavior. In this simulation study, we extend this approach to control and embed not only one but also multiple behaviors into a type of soft robot called a tensegrity robot. The resulting system, consisting of the robot and the environment, is a multistable dynamical system that converges to different attractors from varying initial conditions. Furthermore, attractor analysis reveals that there exist "untrained attractors" in the state space of the system outside the training data. These untrained attractors reflect the intrinsic properties and structures of the tensegrity robot and its interactions with the environment. The impacts of these recent findings in PRC remain unexplored in embodied AI research. We here illustrate their potential to understand various features of embodied cognition that have not been fully addressed to date.
\end{abstract}


\maketitle 

\begin{quotation}
    The behavior of intelligent robots must be both adaptive and diverse. It should be adaptive in the sense that minor perturbations from the environment do not ruin the robot's behavior and diverse in the sense that it can display different behaviors according to the situation. Therefore, intelligent robots may be understood as multistable dynamical systems, which converge to different stable attractors from different initial conditions. Previous studies in the field of reservoir computing have shown ways to design a high-dimensional multistable system using artificial neural networks (ANNs). In this study, we extend this approach to physical reservoir computing using the nonlinear dynamics of soft tensegrity robots as a computational resource instead of ANNs. We propose that this approach, termed \emph{multifunctional physical reservoir computing}, is an effective method of dynamic locomotion control for soft robots.
\end{quotation}

\section{Introduction}

In dynamical systems, attractors are regions in the state space that the system converges to from a range of initial conditions. By definition, state trajectories that belong to an attractor are robust against minor perturbations, as long as they are not pushed out of the basin of attraction. Now, in the context of robotics, the expectation of a reliable robot is that its behavior is, for one, resilient to noise and disturbances from the environment, and two, reproducible despite some slight differences in the state of the robot. For example, a quadruped robot should not lose its balance over a tiny rock on the ground, and a robotic arm should be able to reach the same object from different initial postures. Therefore, it is beneficial to understand, generate, design, or control the behavior of robots as attractors in a dynamical system \cite{beer1995dynamical,warren2006dynamics,khatib1986real,schoner1995dynamics,johnson2008moving,schaal2006dynamic,ijspeert2013dynamical,saveriano2023dynamic,khansari2011learning,taga1991self,tani1999learning,tani2004self,ijspeert2007swimming,steingrube2010self,kuniyoshi2004dynamic,pfeifer2007self,kuniyoshi2019fusing}.

However, another requirement of practical robots is that they should be able to exhibit multiple stable behaviors according to the situation. The situation here can refer to the state of the robot body; depending on how fast a quadruped robot is currently moving forward, it can converge to either a walking or running gait. In addition, the situation can also refer to the environment state; different object positions would require different reaching behavior from a robotic arm. Therefore, robots, or robot-environment systems to be more precise, should be multistable dynamical systems, which are systems with multiple attractors in the state space. Some previous studies leverage multistability, which is innate to the robot-environment system, for exploring and generating the behavior of robots \cite{kuniyoshi2004dynamic,ijspeert2007swimming,steingrube2010self,tani2004self}. Recent works in soft robotics also exploit multistability of the robot's mechanical structure for flexible actuation \cite{chi2022bistable,bhovad2019peristaltic,pagano2017crawling,rothemund2018soft}. However, it is still understudied how multistability can effectively be designed or controlled in the high-dimensional state space of robots.

Recently, a machine learning framework called reservoir computing (RC) has been used to study multistability in high-dimensional artificial neural networks \cite{flynn2021mf,flynn2023seeing,flynn2023theory,o2025confabulation,du2024multi,lu2020invertible,inoue2020designing,kong2024reservoir} (ANNs). Using these methods, it is possible to train recurrent neural networks to reconstruct multiple desired attractors, forcing these networks to become multistable dynamical systems. Therefore, multistability in ANNs can be designed and controlled to some extent, and their dynamical properties, such as basin and bifurcation structures, can be investigated in detail.

At the same time, the extension of RC to physical reservoir computing \cite{nakajima2020physical} (PRC) allows the dynamics of physical systems to be used in computation instead of ANNs. In particular, the sensory and motor capabilities of soft robots can be enhanced with PRC by exploiting the nonlinear physical dynamics that are present in soft components of the robot. When PRC is integrated into the feedback control loop of a soft robot, the computational load required to generate the motor command is largely outsourced to the dynamics of the physical body \cite{nakajima2014exploiting,nakajima2015information,nakajima2018exploiting,hasegawa2024takorobo,zhao2013spine,horii2021physical,bhovad2021physical,wang2023building,tanaka2021flapping,yu2023tapered,sakurai2022durable,hayashi2022online,akashi2024embedding,caluwaerts2013locomotion,caluwaerts2014design,fujita2018environmental}, thereby increasing efficiency in terms of computation and energy. Robot behavior that is produced by these PRC setups is often an attractor of the robot-environment system.

In this paper, we explicitly design multistability in the dynamics of a soft tensegrity robot using PRC in simulation. In particular, this study is the first implementation of multifunctional RC \cite{flynn2021mf,flynn2023seeing,flynn2023theory,o2025confabulation,du2024multi} in soft robot-based PRC, and therefore, we call this approach \emph{multifunctional physical reservoir computing} (MF-PRC). A crucial difference between RC using an ANN and PRC using a soft robot is that the latter is an embodied system, which means that any dynamic trajectory or attractor can be directly (and is inevitably) observed as robot behavior. We also conduct detailed investigations on the multistability of the MF-PRC system and discuss how it is connected to properties that are intrinsic to the physical reservoir, that is, the tensegrity robot. Furthermore, we will argue that byproducts and failures of behavior learning are just as important as its successes for the understanding of embodied intelligence. Experimental results of MF-PRC are presented to support this claim.

The structure of the rest of the paper is as follows. In Sec.~\ref{sec:tensegrity}, we provide an introduction on tensegrity robots. The details of the simulation setup used in the current study are explained in Sec.~\ref{sec:simulation}. We introduce the concept of RC and multifunctional RC in Sec.~\ref{sec:reservoir} and then explain our methods of conducting MF-PRC on tensegrity robots in Sec.~\ref{sec:experiment} and \ref{sec:evolution}. In Sec.~\ref{sec:results}, we present our main findings and then provide discussions on the implications of the results in Sec.~\ref{sec:discussion}. Finally, we conclude the study in Sec.~\ref{sec:conclusion}.

\section{Tensegrity structures}
\label{sec:tensegrity}
In this study, we conduct investigations of physical reservoir computing (PRC) in a simulation of a soft robot comprising tensegrity structures. Tensegrity is a structure composed of rigid and elastic elements, and the entire structure holds its shape by balancing tensile forces. When used as a soft robot, tensegrity structures exhibit high compliance and deformability, allowing various physical interactions and operations in unknown environments \cite{bliss2012central,mirletz2015goal,rieffel2018adaptive,vespignani2018design,baines2020rolling,peng2024unified,wang2019light,zappetti2017bio,shah2022tensegrity}. Previous research has shown that the most popular form of tensegrity robot, which consists of six rigid bars, can exhibit a variety of dynamic locomotion behaviors, such as crawling, rolling, and hopping \cite{terajima2021behavioral,doney2020behavioral}. Furthermore, the six-bar structure can be connected in a modular way to construct a soft robotic arm, which can be used for manipulation tasks including reaching and twisting \cite{kobayashi2023active,kobayashi2023large}. The resilience, versatility, and dynamic nature of soft tensegrity robots make them an interesting topic of study in terms of behavior control and analysis.

In addition, tensegrity structures have some connections to biology. The cytoskeleton that provides mechanical support to and maintains the shapes of cells can be considered to be tensegrity structures \cite{ingber2014tensegrity}. Tensegrities are very responsive to mechanical perturbations, because all components are connected by tensile elements that transmit tensional forces throughout the whole system. This property explains how cells can sense, communicate, and respond to mechanical forces in a process called mechanotransduction. Tensegrities can also be combined in a hierarchical and multimodular manner without losing their stability. Thus, cells are also viewed as hierarchical tensegrity structures, with smaller elements such as the nucleus and the submembranous cytoskeleton forming independent tensegrities within the larger structure.

Information processing through physical computation is another functionality that could be provided by tensegrities. The current study is a step in this direction; however, it is not the first to focus on the computational capabilities of tensegrity robots. In early work, it was proposed that the high degree of dynamic coupling in tensegrity robots allows them to be controlled by simple control inputs, with significantly fewer dimensions than the robot's total degrees of freedom \cite{paul2006morphological}. Another study demonstrated that the network of tension in tensegrity robots acts as a substrate for propagating information, similar to the mechanotransduction process described above \cite{rieffel2010morphological}. Later, Caluwaerts et al. implemented PRC on tensegrity robots, using the dynamics of the robot as a source of nonlinear computation to generate the motor signals necessary for locomotion. This experiment was conducted both in simulation \cite{caluwaerts2013locomotion} and on hardware \cite{caluwaerts2014design}. This work proposed that the control problem of robots can be viewed as a tradeoff between brain (or silicon) computation and body computation. At one end of the spectrum, a classic robot with a stiff body can be controlled by a highly complex controller. At the other end of the spectrum, a soft robot with a highly dynamic and compliant body can be controlled by a simple controller, such as linear feedback, to achieve similar levels of complexity. The control of tensegrity robots using PRC positions itself on the latter end of the spectrum.

\subsection{Simulation settings}
\label{sec:simulation}

We conduct all investigations in the simulation of a six-bar tensegrity robot that is implemented on the physics engine MuJoCo \cite{todorov2012mujoco}. This physics engine is chosen for its support of elastic tendon dynamics and its precision in simulating contacts, making it a suitable platform for investigating the locomotion behaviors of tensegrity robots.

\begin{figure*}
    \includegraphics[]{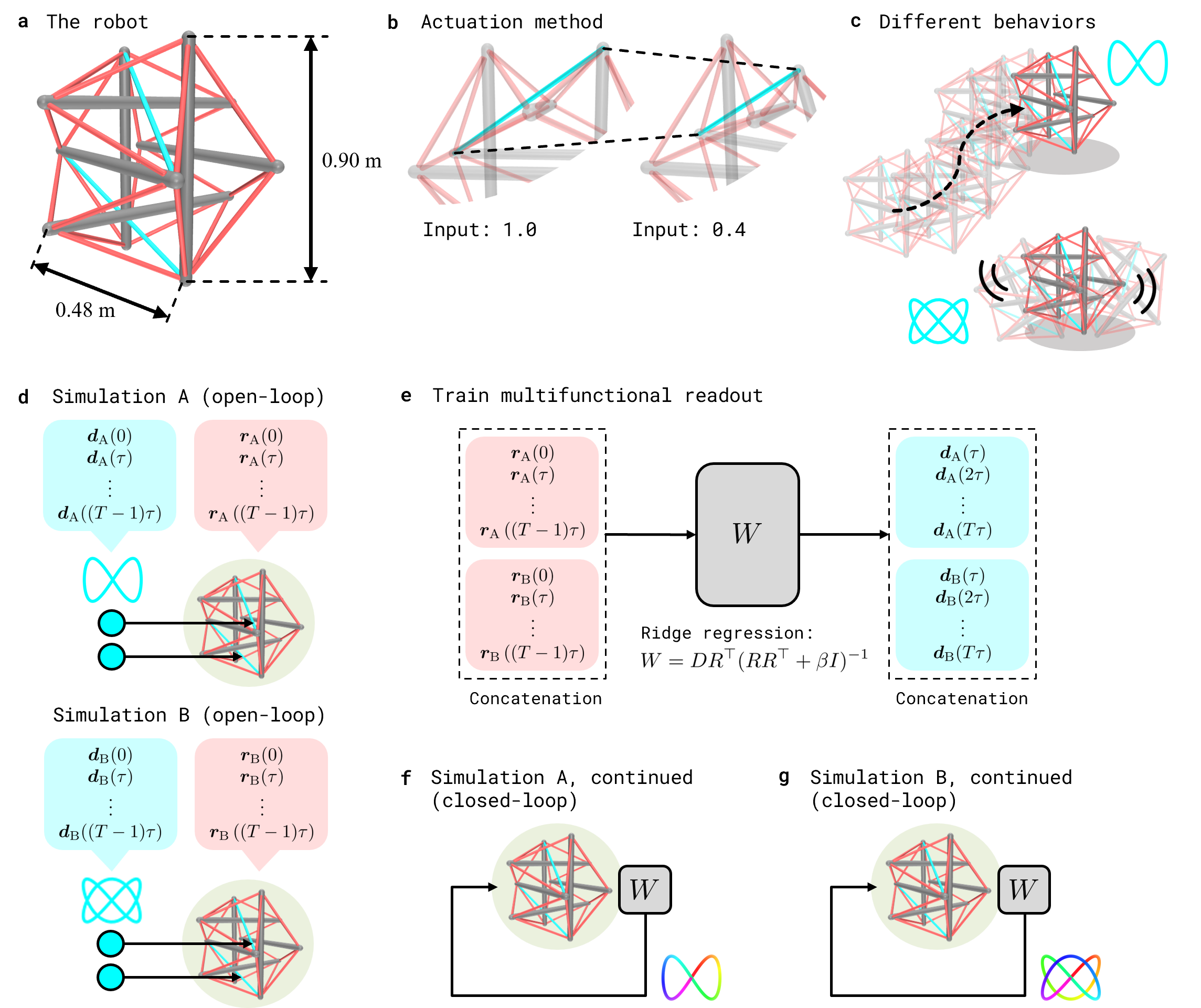}
    \caption{\label{fig:system} Overview of the proposed method. (a) Dimensions of the tensegrity robot at its equilibrium state. (b) Motor signals change the restlength of actuators, which deforms the robot's shape. Each actuator can be controlled independently. (c) Different motor signals, expressed as Lissajous curves, produce different robot behaviors such as forward movement and oscillation. (d) Two separate simulations are conducted for the collection of training data. In Simulation A, the robot is driven by target signal $\bm{d}_\mathrm{A}$. In Simulation B, the robot is driven by target signal $\bm{d}_\mathrm{B}$. The actuators are driven sinusoidally, resulting in Lissajous curves in the input space. (e) The collected reservoir data are concatenated to produce the data matrix $R$. The target signals of the next timestep are concatenated to produce the target matrix $D$. The multifunctional readout $W$ is obtained via ridge regression on $R$ and $D$. (f) The closed-loop phase continues from where the robot was left in Simulation A of the open-loop phase. If training is successful, the robot replicates the same behavior as in the open-loop phase. (g) The closed-loop phase is also continued for Simulation B. The same $W$ can be used to produce different robot behaviors in Simulations A and B.}
\end{figure*}

As shown in Fig.~\ref{fig:system}(a), the robot consists of six rigid bars of equal length, connected by 24 passive tendons and two actuated tendons (actuators). Each rigid bar is a cylindrical object with an evenly distributed mass, and the force exerted by each tendon is calculated using the following equation:
\begin{equation}
    \label{eq:tendon}
    F_\mathrm{s} = - k(x - x_\mathrm{r}) - c \dot{x},
\end{equation}
where $F_\mathrm{s}$ is the tendon force, $k$ is the tendon stiffness, $x$ is the current tendon length, $x_\mathrm{r}$ is the tendon restlength, and $c$ is the damping coefficient. Eq.~(\ref{eq:tendon}) is expressed in tendon coordinates, with a positive sign for compressive forces and a negative sign for tensile forces. Parameters $k$ and $c$ are set to be the same for all tendons, including passive tendons and actuators.

If we remove the actuators and disable all external forces such as gravity, the tensegrity robot relaxes to a passive equilibrium state. Because the arrangement of passive tendons is symmetrical with respect to the robot's center of mass, the equilibrium state is also symmetrical, forming Jessen's orthogonal icosahedron \cite{gorkavyy2016model}. The actuators (exerting zero force at this point) are then attached as shown in Fig.~\ref{fig:system}(a). Due to the symmetry of the structure, the lengths of the actuators at this equilibrium state are the same for both actuators; we denote this length as $L$. Note, however, that the placement of the actuators is not symmetrical with respect to the robot's center of mass. In preliminary experiments, we found that the robot tends to exhibit a wider variety of behaviors when its symmetry is broken.

The actuators are activated by motor signals $u_1(t), \, u_2(t) > 0$ that change their restlengths independently according to the following equation:
\begin{align}
    x_{\mathrm{r}, 1} (t) & = L u_1(t) \\
    x_{\mathrm{r}, 2} (t) & = L u_2(t),
\end{align}
causing the robot to deform as in Fig.~\ref{fig:system}(b). In the current experiment setting, the values of $u_1(t)$ and $u_2(t)$ can be larger than 1, which means that the actuators' restlengths can be longer than their original values of $L$. Hereafter, we also refer to motor signals as the system input, $\bm{u}(t) = [u_1(t), \, u_2(t)]^\top$. Constantly changing inputs cause the robot to deform continuously, resulting in various dynamic movements. A previous study has shown that, under similar simulation settings, the tensegrity robot can exhibit stable locomotion, such as crawling and rolling, in-place movements similar to shaking and hopping, and chaotic behaviors \cite{terajima2021behavioral}.

The simulation timestep of the physics engine is set to 1~ms. Further details of the robot, the simulation, and their parameter values are provided in Appendix~\ref{app:physics}.

\section{Multifunctional physical reservoir computing}
\label{sec:mfprc}

\subsection{Multifunctional reservoir computing}
\label{sec:reservoir}
Reservoir computing \cite{jaeger2001echo,maass2002real,verstraeten2007experimental} (RC) is a scheme of machine learning that is most commonly implemented using echo state networks \cite{jaeger2001echo} (ESNs), a type of artificial neural network (ANN) with recurrent connections. An ESN consists of three layers: the input layer, the reservoir layer, and the readout layer. The reservoir layer is a high-dimensional network with random recurrent connections, and therefore, it is represented by a high-dimensional state space. The characteristic of RC is that the internal weights of the network, which are weights in the input and reservoir layers, are fixed throughout training. As long as the reservoir meets certain conditions, it is sufficient to train only the readout layer to obtain the desired input-output relationship and perform a wide range of computational tasks. This contrasts with the traditional machine learning methods of ANNs, where all of the internal weights are trained using, for example, backpropagation.

In RC, the role of the reservoir layer with fixed weights is to provide short-term memory and nonlinear transformations of the input. Learning, or the update of weights, is executed solely by the readout layer. Therefore, it is not necessary for the reservoir to be an ANN; it can be any arbitrary dynamical system that possesses short-term memory and nonlinearity. In other words, the dynamics simulated by the ANN in a conventional RC system can be replaced by physical dynamics. This is the basis of physical reservoir computing \cite{nakajima2020physical} (PRC). 
According to previous studies, a bucket of water \cite{fernando2003pattern}, biological neuronal cultures \cite{dranias2013short,ishida2023quantification,cai2023brain}, photonic \cite{van2017advances,zhang2025mf}, spintronic \cite{torrejon2017neuromorphic,furuta2018macromagnetic,tsunegi2019physical,akashi2020input,akashi2022coupled,tsunegi2023information,yamaguchi2023computational}, and quantum devices \cite{fujii2017harnessing,nakajima2019boosting,ghosh2021quantum,fujii2021quantum,tran2021learning,kubota2023temporal} can all be effectively used as a physical reservoir. Another type of physical reservoir that deserves interest is the dynamics of mechanical systems, especially soft robots. The information processing capabilities of mass-damper spring systems \cite{hauser2011towards,hauser2012role}, silicone tentacles resembling octopus arms \cite{nakajima2014exploiting,nakajima2015information,nakajima2018exploiting,hasegawa2024takorobo}, quadruped robots with soft spines \cite{zhao2013spine}, fish robots \cite{horii2021physical}, origami robots \cite{bhovad2021physical,wang2023building}, winged robots \cite{tanaka2021flapping}, whiskered robots \cite{yu2023tapered}, and artificial muscles \cite{sakurai2022durable,hayashi2022online,akashi2024embedding} have been investigated in the context of PRC.
As introduced previously, tensegrity structures are also used as physical reservoirs \cite{caluwaerts2013locomotion,caluwaerts2014design,fujita2018environmental}.

One important computational task in RC and PRC is attractor reconstruction, where the objective is to emulate an attractor of another dynamical system (usually called the target attractor) with the dynamics of the reservoir. In this task, the readout layer is trained using observations collected from the target attractor, so that the RC system's input-output relationship approximates the evolution of observations from one timestep to the next. The training data required to learn this relationship are collected in the \emph{open-loop} configuration of RC, where the reservoir is driven externally by the observations. After training, the output of the RC system is fed back to the reservoir as the next input, allowing the system to evolve autonomously and reproduce the target attractor without being driven by the observations. This is called the \emph{closed-loop} configuration of RC. When the training of attractor reconstruction is successful, it is known that RC systems can not only make a short-term forecast of the chaotic target attractor, but also reproduce its long-term ergodic properties \cite{lu2018attractor}. This capability is also confirmed in PRC systems \cite{akashi2024embedding}.

Attractor reconstruction can be extended to memorize multiple target attractors simultaneously. Usually, multitasking in RC is achieved by preparing a separate readout layer for each task, while re-using the same reservoir layer. However, it is also possible to learn and reproduce multiple target attractors using the same reservoir and readout layers. This is called multifunctional reservoir computing \cite{flynn2021mf,flynn2023seeing,flynn2023theory,o2025confabulation,du2024multi} (MF-RC). Here, \emph{multifunctionality} means that the RC system is multistable; the system possesses multiple attractors in its state space and converges to different attractors from different initial conditions. In MF-RC, the training data are a concatenation of observations from multiple target attractors, and readout weights can be obtained either by batch training \cite{flynn2021mf} or online training \cite{lu2020invertible}. Alternatively, the observations of each target attractor can be provided along with a unique label, so that it is easier to select which attractor to reconstruct during inference (i.e., after training), akin to associative memory \cite{o2025confabulation,kong2024reservoir,du2024multi,inoue2020designing}. This label can be considered an external cue that switches the initial conditions of the reservoir, allowing the system to converge to the selected attractor.

Furthermore, analyses of the basin and bifurcation structures of MF-RC reveal that within the state space of the reservoir, there exist attractors that are different from the targets given during training \cite{flynn2021mf,flynn2023seeing,flynn2023theory,o2025confabulation,du2024multi,morra2023mf}. To distinguish these emergent attractors from the reconstructed target attractors, they can be called \emph{untrained attractors} \cite{flynn2021mf,flynn2023seeing}. At first glance, untrained attractors may be perceived as undesirable behavior of the MF-RC system. However, they can also be exploited as a method of generating new chaotic dynamics \cite{du2024multi,carroll2022creating,kabayama2024designing}.

In the following sections, we describe our extension of MF-RC to a tensegrity reservoir system, namely, \emph{multifunctional physical reservoir computing} (MF-PRC). When MF-PRC successfully learns multiple target attractors, the tensegrity robot produces different behaviors from different initial conditions.

\subsection{Procedure for MF-PRC on tensegrity robots}
\label{sec:experiment}
The procedure for conducting MF-PRC on a tensegrity robot resembles that of MF-RC \cite{flynn2021mf}, except that the ESN is replaced by the robot as a physical reservoir. Similarly, it consists of three stages: open-loop simulation, training of the multifunctional readout, and closed-loop simulation. The overview of the procedure is shown in Fig.~\ref{fig:system}(d)-(g).

In the open-loop simulation stage, we conduct two separate simulations, \emph{Simulation A} and \emph{Simulation B}. Both simulations begin from the same state, in which the tensegrity robot sits still on the ground. In each simulation, the tensegrity robot is driven by different target motor signals, $\bm{d}_\mathrm{A}(t), \;\bm{d}_\mathrm{B}(t)\in \mathbb{R}^2$, and produces distinct behaviors, as illustrated by Fig.~\ref{fig:system}(c). The motor signals are updated every $\tau = 10~\mathrm{ms}$, which is defined as the reservoir timestep. (Note that this timestep is different from the simulation timestep explained in Sec.~\ref{sec:simulation}. Hereafter, we always refer to the reservoir timestep when we simply mention a "timestep" or "step.") The simulations are continued for a total of $T = 20,000$ timesteps, and at each timestep, the 48-dimensional vector containing the lengths and velocities of the passive tendons is recorded as the reservoir measurement, $\bm{r}_\mathrm{A}(t), \;\bm{r}_\mathrm{B}(t) \in \mathbb{R}^{48}$. In contrast, the state of the full dynamical system, $\bm{s}_\mathrm{A}(t)$ or $\bm{s}_\mathrm{B}(t)$, is represented by the positions and velocities of the rigid bars, following the convention of the physics simulator MuJoCo. The reservoir measurement is an observation of the system state. At the end of the open-loop simulations, their final states are recorded as $\bm{s}_\mathrm{A}^\mathrm{I} = \bm{s}_\mathrm{A}(T\tau)$ and $\bm{s}_\mathrm{B}^\mathrm{I} = \bm{s}_\mathrm{B}(T\tau)$ so that later simulations can continue from these states. An illustration of the open-loop simulation stage is provided in Fig.~\ref{fig:system}(d).

In the next stage, we use the data collected from the open-loop simulations to train the multifunctional readout, as illustrated in Fig.~\ref{fig:system}(e). The recorded reservoir measurements from Simulations A and B are concatenated to produce the following reservoir data matrix $R \in \mathbb{R}^{48 \times 2T}$,
\begin{equation*}
    R = \left[ \bm{r}_\mathrm{A}(0), \dots, \bm{r}_\mathrm{A}((T-1)\tau), \bm{r}_\mathrm{B}(0), \dots, \bm{r}_\mathrm{B}((T-1)\tau) \right],
\end{equation*}
and target motor signals corresponding to the next timestep are concatenated to produce the target data matrix $D \in \mathbb{R}^{2 \times 2T}$,
\begin{equation*}
    D = \left[ \bm{d}_\mathrm{A}(\tau), \dots, \bm{d}_\mathrm{A}(T\tau), \bm{d}_\mathrm{B}(\tau), \dots, \bm{d}_\mathrm{B}(T\tau) \right].
\end{equation*}
Using this procedure, the training data from both simulations are combined. Then, the weights $W \in \mathbb{R}^{2 \times 48}$ of the readout layer are obtained by ridge regression on $R$ and $D$, given by
\begin{equation}
    W = D R^\top (R R^\top + \beta I)^{-1},
\end{equation}
where $\beta$ is the regularization parameter and $I \in \mathbb{R}^{48 \times 48}$ is the identity matrix. Ridge regression penalizes large elements in $W$, and therefore, it is effective in preventing overfitting and increasing stability in the following closed-loop simulation stage. In our training experiments, the value of $\beta = 0.01$ is used.

After training, the output of the readout layer is fed back to the robot as next motor commands, forming a closed-loop system. This system is an autonomous dynamical system that can be expressed with the following equations:
\begin{align}
    \bm{s}(t + \tau) & = f(\bm{s}(t), \bm{u}(t)), \\
    \bm{r}(t) & = g(\bm{s}(t)) \\
    \bm{u}(t+\tau) & = W \bm{r}(t),
\end{align}
where $f$ is the physics engine (including actuation) and $g$ is a function that extracts reservoir measurements from the simulation state. As illustrated in Fig.~\ref{fig:system}(f) and (g), we run two separate closed-loop simulations that continue the open-loop Simulations A and B; this is equivalent to initializing the closed-loop simulations with states $\bm{s}_\mathrm{A}^\mathrm{I}$ and $\bm{s}_\mathrm{B}^\mathrm{I}$, respectively. If training is successful, the robot should reproduce the same long-term behaviors (attractors) as in the open-loop simulation stage. Significantly, the same readout $W$ is used to approximate both target motor signals that were given in the open-loop stage, $\bm{d}_\mathrm{A}(t)$ and $\bm{d}_\mathrm{B}(t)$. This is possible because the trajectories of the reservoir measurements are well separated for the two simulations. As a result, the whole system becomes a multistable dynamical system, with different attractors for different initial conditions (such as $\bm{s}_\mathrm{A}^\mathrm{I}$ and $\bm{s}_\mathrm{B}^\mathrm{I}$).


In principle, the aforementioned procedure of MF-PRC can be applied to any combination of target motor signals and extended to any number of targets. However, the objective of this study is to demonstrate that MF-PRC can be used to control and generate interesting behaviors for soft robots, rather than to validate the universality of our approach. To achieve this objective, we optimize a pair of target motor signals using an evolutionary algorithm. Details of this optimization process are provided in the next section.

\begin{figure}
    \includegraphics[]{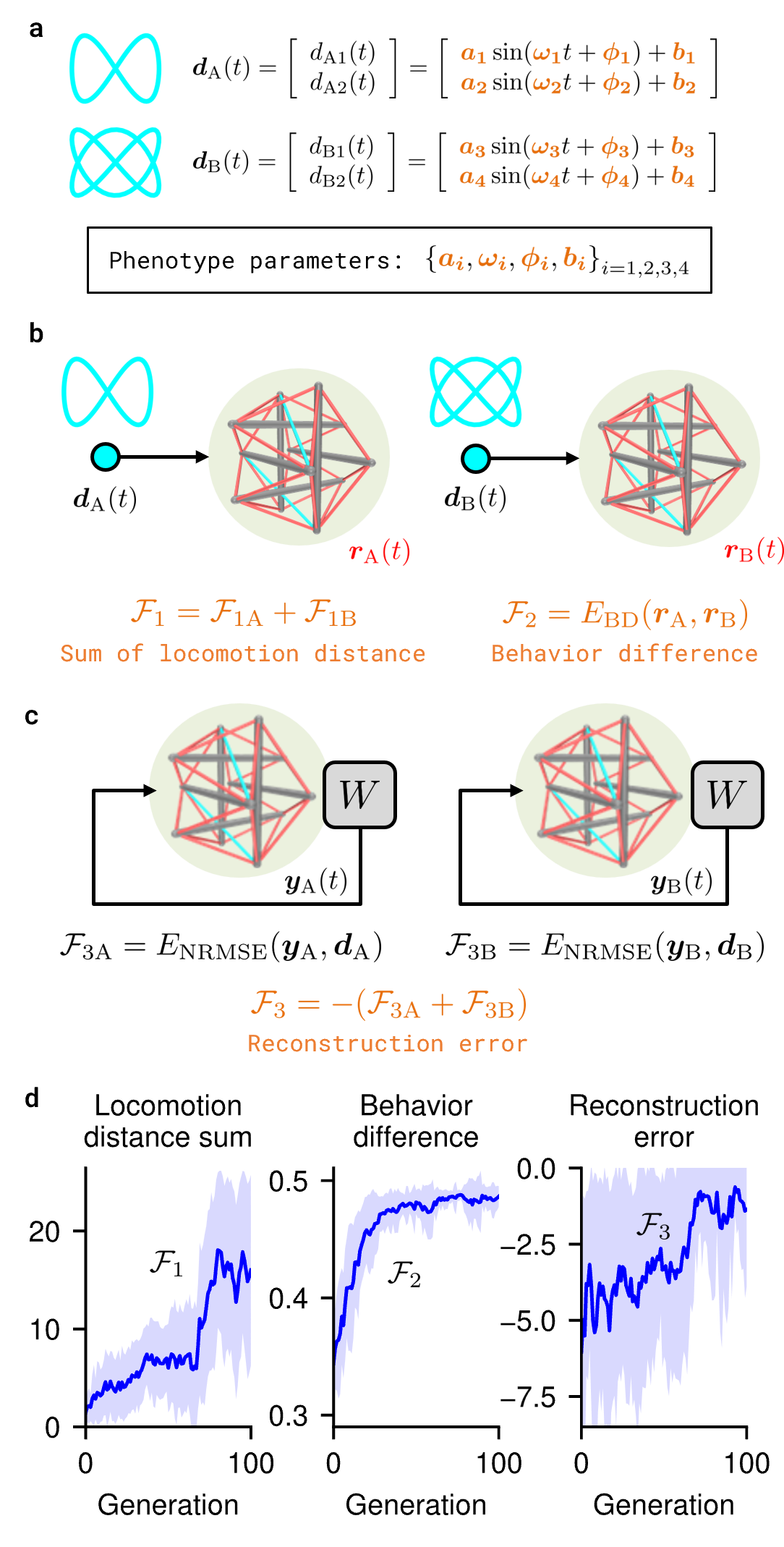}
    \caption{\label{fig:evolution} Optimization of target signals with a genetic algorithm. (a) Each phenotype is a pair of target signals in the form of Lissajous curves, expressed by 16 parameters. (b) The target signals are used to drive the tensegrity robot in open-loop. The sum of locomotion distance is used as the first fitness objective. The behavior difference, defined as the difference in reservoir measurements, is used as the second fitness objective. The final state of the open-loop simulations are recorded as $\bm{s}_\mathrm{A}^\mathrm{I}$ and $\bm{s}_\mathrm{B}^\mathrm{I}$. (c) The robot is controlled in closed-loop using the multifunctional readout, obtained by training. The closed-loop simulations are initialized with states $\bm{s}_\mathrm{A}^\mathrm{I}$ and $\bm{s}_\mathrm{B}^\mathrm{I}$. The sum of reconstruction error, defined as the normalized root mean squared error between the output and the target signals, is used as the third fitness objective. (d) The fitness curve for each objective over 100 generations. From the left, the objective is the sum of locomotion distance, behavior difference, and reconstruction error. Solid lines and shaded regions represent the mean and standard deviation, respectively, of each objective within the generation. As intended, all three objectives are maximized by the genetic algorithm.}
\end{figure}

\subsection{Evolutionary optimization of target input}
\label{sec:evolution}
The objective of this evolutionary optimization is to obtain a good pair of target motor signals, $\bm{d}_\mathrm{A}(t)$ and $\bm{d}_\mathrm{B}(t)$, which produces interesting robot behaviors and can successfully be learned by MF-PRC. Here, this objective is decomposed into the following three conditions. First, the robot should exhibit locomotion behaviors that travel a long distance ($\mathcal{F}_1$). Second, the two motor signals should produce qualitatively different behaviors, so that the MF-PRC system can achieve multifunctionality ($\mathcal{F}_2$). Third, it must be possible for the tensegrity physical reservoir to learn these target motor signals ($\mathcal{F}_3$). Each of these conditions is quantified as a separate fitness objective, and the target motor signals are optimized using a multi-objective genetic algorithm that maximizes these objectives simultaneously. The overall process is illustrated in Fig.~\ref{fig:evolution}, and its details are as follows.

For simplicity, the target motor signals are represented as Lissajous curves in the following form, as in Fig.~\ref{fig:evolution}(a):
\begin{align}
    \label{eq:targetA}
    \bm{d}_\mathrm{A}(t) &=
    \left[
        \begin{array}{r}
            a_1 \sin (\omega_1 t + \phi_1) + b_1 \\
            a_2 \sin (\omega_2 t + \phi_2) + b_2
        \end{array}
    \right] \\
    \label{eq:targetB}
    \bm{d}_\mathrm{B}(t) &=
    \left[
        \begin{array}{r}
            a_3 \sin (\omega_3 t + \phi_3) + b_3 \\
            a_4 \sin (\omega_4 t + \phi_4) + b_4
        \end{array}
    \right],
\end{align}
with the tuple of 16 parameters $\left\{ a_i, \omega_i, \phi_i, b_i \right\} (i=1,2,3,4)$ representing each phenotype (individual) in the genetic algorithm. This experiment choice notably limits the range of possible targets to periodic motor signals; however, it is favorable for the objective of obtaining locomotion behaviors. Moreover, periodic motor signals do not necessarily indicate periodic robot behaviors \cite{terajima2021behavioral}, so the diversity of possible robot behaviors is maintained. The phenotype parameters are sampled from uniform distributions within reasonable value ranges.

For each phenotype (pair of target signals), the procedure of MF-PRC is conducted as described in Sec.~\ref{sec:experiment}. As shown in Fig.~\ref{fig:evolution}(b), the first and second fitness objectives are calculated during the open-loop simulations, in which the robot is externally driven by the target motor signals. The first fitness objective, $\mathcal{F}_1$, is defined as the sum of the locomotion distance displayed by the robot in the open-loop Simulations A and B. More specifically, the trajectory of the robot's center of mass in global coordinates is recorded during the final 5,000 steps (50 seconds) of each simulation. The Euclidean distance between the start and end points of the trajectory is defined as the locomotion distance.

The second fitness objective, $\mathcal{F}_2$, quantifies the difference in robot behaviors observed in Simulations A and B. Without this objective, the two target signals of a single phenotype may converge to the same signal during the evolution, which is an easy solution for maximizing the other two objectives. This clearly invalidates the concept of multifunctionality. To prevent this, the behavior difference is computed from the time series of the reservoir measurements $\bm{r}_\mathrm{A}(t)$ and $\bm{r}_\mathrm{B}(t)$ during the final 5,000 steps of the open-loop simulations. First, both $\bm{r}_\mathrm{A}(t)$ and $\bm{r}_\mathrm{B}(t)$ are normalized to have zero mean and unit variance, denoted as $\bm{z}_\mathrm{A}(t)$ and $\bm{z}_\mathrm{B}(t)$. Then, the normalized root mean squared error (NRMSE) between $\bm{z}_\mathrm{A}(t)$ and the time-shifted series $\bm{z}_\mathrm{B}(t + \Delta t)$ is computed. Finally, the behavior difference $\mathcal{F}_2$ is defined as the minimum NRMSE over time shifts $\Delta t$ within a certain range. The maximization of this fitness objective encourages the robot to produce reservoir measurements with different temporal patterns for targets $\bm{d}_\mathrm{A}(t)$ and $\bm{d}_\mathrm{B}(t)$, which indicate qualitatively distinct behaviors.

The final fitness objective is calculated after the training and closed-loop simulation stages of the MF-PRC procedure, as shown in Fig.~\ref{fig:evolution}(c). The closed-loop simulations, which are continuations of open-loop Simulations A and B, are conducted for 20,000 steps each, and the final 5,000 steps are used for evaluation. In these simulations, the outputs of the readout layer are denoted by $\bm{y}_\mathrm{A}(t)$ and $\bm{y}_\mathrm{B}(t)$. The minimum NRMSE considering time shifts between $\bm{y}_\mathrm{A}(t)$ and $\bm{d}_\mathrm{A}(t)$ is computed as the reconstruction error for Simulation A, denoted by $\mathcal{F}_{3\mathrm{A}}$. Similarly, the reconstruction error for Simulation B is denoted by $\mathcal{F}_{3\mathrm{B}}$. Then, the third fitness objective is defined by $\mathcal{F}_3 = -(\mathcal{F}_{3\mathrm{A}} + \mathcal{F}_{3\mathrm{B}})$. The maximization of this objective using a genetic algorithm is convenient because the computational capability of tensegrities in PRC is not universal. In other words, not all target signals (and their combinations) can be learned by the tensegrity reservoir. Including this fitness objective constrains the search space of the genetic algorithm to the class of target signals that can be learned by the current tensegrity robot.

The genetic algorithm used to optimize $\mathcal{F}_1, \mathcal{F}_2, \mathcal{F}_3$ simultaneously is the NSGA-II algorithm \cite{deb2002fast}, a common choice for multi-objective optimization problems. It is run for 100 generations, with a population size of 64. Other experiment details are provided in Appendix~\ref{app:evolution}.

\section{Results}
\label{sec:results}

In this section, we present the results of our experiments. First, we briefly describe the course of the evolutionary optimization of target motor signals in Sec.~\ref{sec:result_evo}. We then pick up a pair of targets from the evolution and demonstrate a successful case of MF-PRC in Sec.~\ref{sec:result_recon} as the main result of this paper. Additionally, in Sec.~\ref{sec:result_untrained}, we find many untrained attractors that emerge as byproducts of the MF-PRC training. We follow up these results with an analysis on attractor basins in Sec.~\ref{sec:result_basin} and response of the closed-loop system to changes in body parameters in Sec.~\ref{sec:result_adapt}. Finally, we investigate the parameter regions that allow successful MF-PRC in Sec.~\ref{sec:result_bif}.

\begin{figure*}
    \includegraphics[]{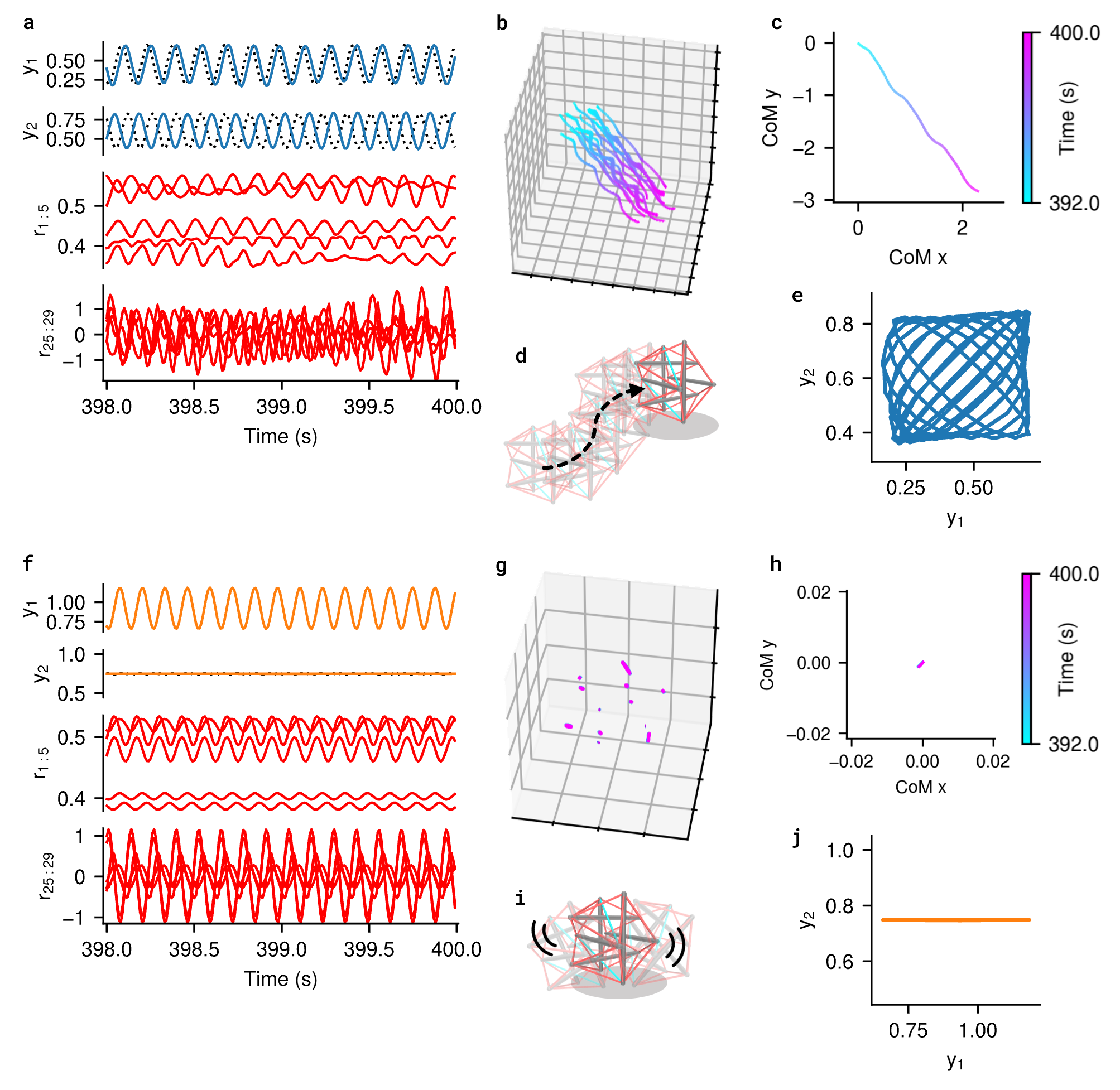}
    \caption{\label{fig:dynamics} The system behavior of a successfully trained multifunctional physical reservoir. (a) Time series of the output $(y_1, y_2)$ and the reservoir measurements $\bm{r}(t)$ during the closed-loop phase of Simulation A. The target signal $\bm{d}_\mathrm{A}(t)$ is represented by dotted lines. (b) Global behavior trajectory of the robot in Simulation A. Each line represents the trajectory of a marker, which is attached to the end of a rigid bar that comprises the robot. The grid spacing is 0.5~m. (c) Trajectory of the robot's center of mass from the bird's-eye view. (d) An illustration of the robot behavior. (e) Outputs of the physical reservoir, plotted in the $(y_1, y_2)$ plane. Subfigures (f)--(j) show the results for Simulation B. Overall, the system successfully replicates a 2-D Lissajous curve and displays forward crawling locomotion in Simulation A. The same system successfully replicates a 1-D periodic signal and displays oscillatory behavior in Simulation B.}
\end{figure*}

\subsection{Process of evolution}
\label{sec:result_evo}
Fig.~\ref{fig:evolution}(d) shows the mean and standard deviation of the fitness objectives at each generation of the evolutionary optimization. All fitness values follow an increasing trend, indicating that the genetic algorithm is successful at finding target signal pairs that satisfy the fitness objectives. As the evolution proceeds, the robot behaviors produced by the two target signals become more qualitatively different, and their locomotion distances increase. In addition, the decrease in reconstruction error indicates that the tensegrity physical reservoir can learn the target signals more accurately. 

On close inspection of the optimization process, however, we find that the distribution of robot behaviors within each generation is different at various stages of the evolution. The collection of robot behaviors \emph{within each generation} is most diverse in the early stages of the evolution (roughly up to Generation 20), due to high randomness of the phenotype parameters. Conversely, the robot behaviors produced by the two target signals \emph{within each phenotype} are not necessarily distinct, as indicated by the low value of $ \mathcal{F}_2 $. Furthermore, many of them have short locomotion distances and are not learned successfully by PRC. In a few cases, the closed-loop PRC system is unstable and causes the physics simulation to explode.

In the middle stages of evolution (roughly up to Generation 65), the value of $ \mathcal{F}_2 $ is maximized, and the behaviors produced by the two target signals become more distinct. Despite the improvement in the reconstruction error, MF-PRC training is still not successful in some phenotypes, resulting in instabilities or the imperfect reconstruction of target signals. In such failure cases, the closed-loop system cannot reproduce the behaviors observed in the open-loop simulations, and it always converges to some undesired attractors. (However, we discuss the possibility of exploiting failures in Sec.~\ref{sec:discussion}.) During this middle stage, the number of phenotypes that display long locomotion distances increases gradually.

The locomotion distance and reconstruction error greatly improve in the final stages of evolution (after Generation 65), and the robot can successfully learn locomotion behavior with MF-PRC. However, we also notice the convergence of phenotype parameters to similar values, causing phenotypes within each generation to produce similar behavior patterns. Notably, in our current experiment, many phenotypes from later generations followed a similar pattern where target $ \bm{d}_\mathrm{A}(t) $ produces highly efficient forward locomotion and target $ \bm{d}_\mathrm{B}(t) $ produces small oscillatory movements on the spot (no locomotion). This specific pattern is likely one of the optimal solutions for the given fitness objectives, because it solves the tradeoff between the locomotion distance and the behavior difference. The improvement in reconstruction error suggests that the learned behaviors are stable.

In the next section, we pick up a phenotype that follows this pattern as a representative case of successful MF-PRC. Many phenotypes taken from the final stages of evolution show similar trends. Other interesting cases can be found especially around Generation 65, where the balance between optimality and behavioral diversity is maintained.

\begin{table}
    \caption{\label{tab:phenotype} Phenotype parameter values.}
    \begin{ruledtabular}
    \begin{tabular}{cccc}
        Parameter & Value & Parameter & Value \\
        \hline
        $a_1$ & 0.2563 & $\phi_1$ & 0.4945 \\
        $a_2$ & 0.2264 & $\phi_2$ & 4.214 \\
        $a_3$ & 0.2626 & $\phi_3$ & 3.666 \\
        $a_4$ & 0.01001 & $\phi_4$ & 0.01529 \\
        \hline
        $\omega_1$ & 42.73 & $b_1$ & 0.4467 \\
        $\omega_2$ & 45.24 & $b_2$ & 0.6040 \\
        $\omega_3$ & 48.82 & $b_3$ & 0.9200 \\
        $\omega_4$ & 103.7 & $b_4$ & 0.7480 \\
    \end{tabular}
    \end{ruledtabular}
\end{table}

\begin{figure}
    \includegraphics[]{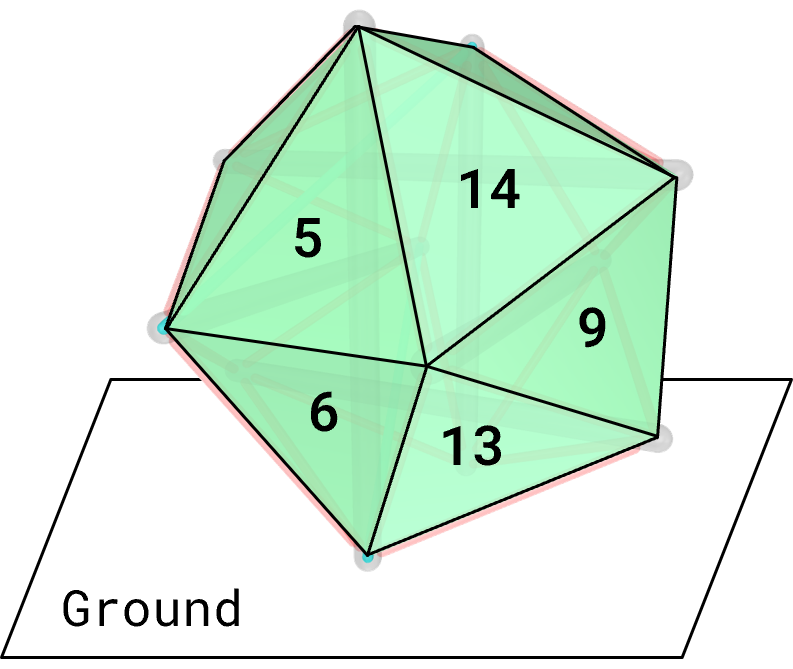}
    \caption{\label{fig:icosa} Viewing the tensegrity robot as an icosahedron. Each face of the icosahedron is labeled with a number from \{1\} to \{20\}. The resulting behavior of the robot can be characterized by the face number that is touching the ground.}
\end{figure}

\subsection{Reconstruction of multiple attractors with MF-PRC}
\label{sec:result_recon}
In this section, we present a successful case of MF-PRC, taken from Generation 73 of the genetic algorithm described above. The phenotype parameter values for this specific case are listed in Table~\ref{tab:phenotype}. A summary of reconstructed outputs, reservoir measurements, and robot behaviors in the closed-loop stage is shown in Fig.~\ref{fig:dynamics}.

Fig.~\ref{fig:dynamics}(a)-(e) show results for closed-loop continuations of Simulation A, and Fig.~\ref{fig:dynamics}(f)-(j) show results for closed-loop continuations of Simulation B. In other words, the initial conditions (IC) of the closed-loop simulations are $\bm{s}_\mathrm{A}^\mathrm{I}$ and $\bm{s}_\mathrm{B}^\mathrm{I}$, respectively. Both simulations use the same trained multifunctional readout $W$, so the only difference in experiment condition is the IC. In the top rows of Fig.~\ref{fig:dynamics}(a) and (f), the output of the physical reservoir $\bm{y}(t) = [y_1(t), \, y_2(t)]^\top$ is plotted (in solid lines) along with the target signals (in dotted lines). There is a phase difference between the output and the target. Nonetheless, the target is successfully reconstructed. The outputs are also plotted on the $(y_1, y_2)$ plane in Fig.~\ref{fig:dynamics}(e) and (j), revealing the shape of the Lissajous curves. In Simulation B, the amplitude of $y_2$ is close to zero, and the output is essentially a one-dimensional sinusoidal curve (rather than a two-dimensional Lissajous curve). This results from target optimization by the genetic algorithm and is not a failure of MF-PRC. Other rows of Fig.~\ref{fig:dynamics}(a) and (f) show samples of reservoir measurements $\bm{r}(t)$, which are passive tendon lengths ($r_1$ to $r_{24}$) and passive tendon velocities ($r_{25}$ to $r_{48}$). The dynamics are quasi-periodic in Simulation A (due to an irrational period ratio between $y_1$ and $y_2$) and periodic in Simulation B.

Robot behaviors produced during the closed-loop stages are qualitatively different in Simulations A and B. Fig.~\ref{fig:dynamics}(b) and (g) show trajectories in global coordinates of the robot's 12 nodes, where each node is defined as the tip of a rigid bar. Similarly, Fig.~\ref{fig:dynamics}(c) and (h) show trajectories of the robot's center of mass from a bird's-eye view. The behavior in Simulation A can best be described as a rapid crawling motion. There is little vertical movement, so it also looks as if the robot is smoothly sliding on the ground. The robot advances with a slightly winding course, and its forward locomotion speed is 0.465 m/s, or roughly 31 body lengths per minute (BL/min). When compared with existing hardware implementations, this speed is faster than most rolling (e.g., 1.15 BL/min \cite{wang2019light}) and crawling (e.g., 11.5 BL/min \cite{zappetti2017bio}) tensegrity robots, but slower than some vibrating (e.g., 69 BL/min \cite{rieffel2018adaptive}) robots. In contrast, in Simulation B, the robot only displays small oscillatory movements, and it does not locomote forward. For clarity, an illustration of each behavior is provided in Fig.~\ref{fig:dynamics}(d) and (i).

These closed-loop behaviors are reproduced even when the simulations are run from slightly different initial conditions, or when small external forces are applied to the robot. Therefore, the two behaviors described in the previous paragraph are attractors of a dynamical system, composed of the robot body, the environment, and the trained readout. Robustness against perturbations is a natural dynamic property of the system, and a carefully designed recovery plan is not necessary. This point is a clear advantage of generating robot behaviors as attractors of a dynamical system, which has been emphasized repeatedly in the literature \cite{beer1995dynamical,warren2006dynamics,khatib1986real,schoner1995dynamics,johnson2008moving,schaal2006dynamic,ijspeert2013dynamical,saveriano2023dynamic,khansari2011learning,taga1991self,tani1999learning,tani2004self,ijspeert2007swimming,steingrube2010self,kuniyoshi2004dynamic,pfeifer2007self,kuniyoshi2019fusing}. However, many existing works focus on learning or designing a single attractor \cite{khatib1986real,schaal2006dynamic,ijspeert2013dynamical}. Extensions or other works that demonstrate multiple dynamic behaviors of robots often rely on the empirical design of the robot's morphology and control policy \cite{steingrube2010self,kuniyoshi2004dynamic}, or they introduce additional parameters or weights that can be manipulated to change behaviors \cite{ijspeert2007swimming,tani2004self,saveriano2023dynamic}. However, our approach of MF-PRC offers a simple way of designing multiple behaviors that can be switched based on system states (i.e., ICs) alone, without the need for additional parameters or weights. More will be discussed in Sec.~\ref{sec:discussion}.

\begin{figure*}
    \includegraphics[height=8.25in]{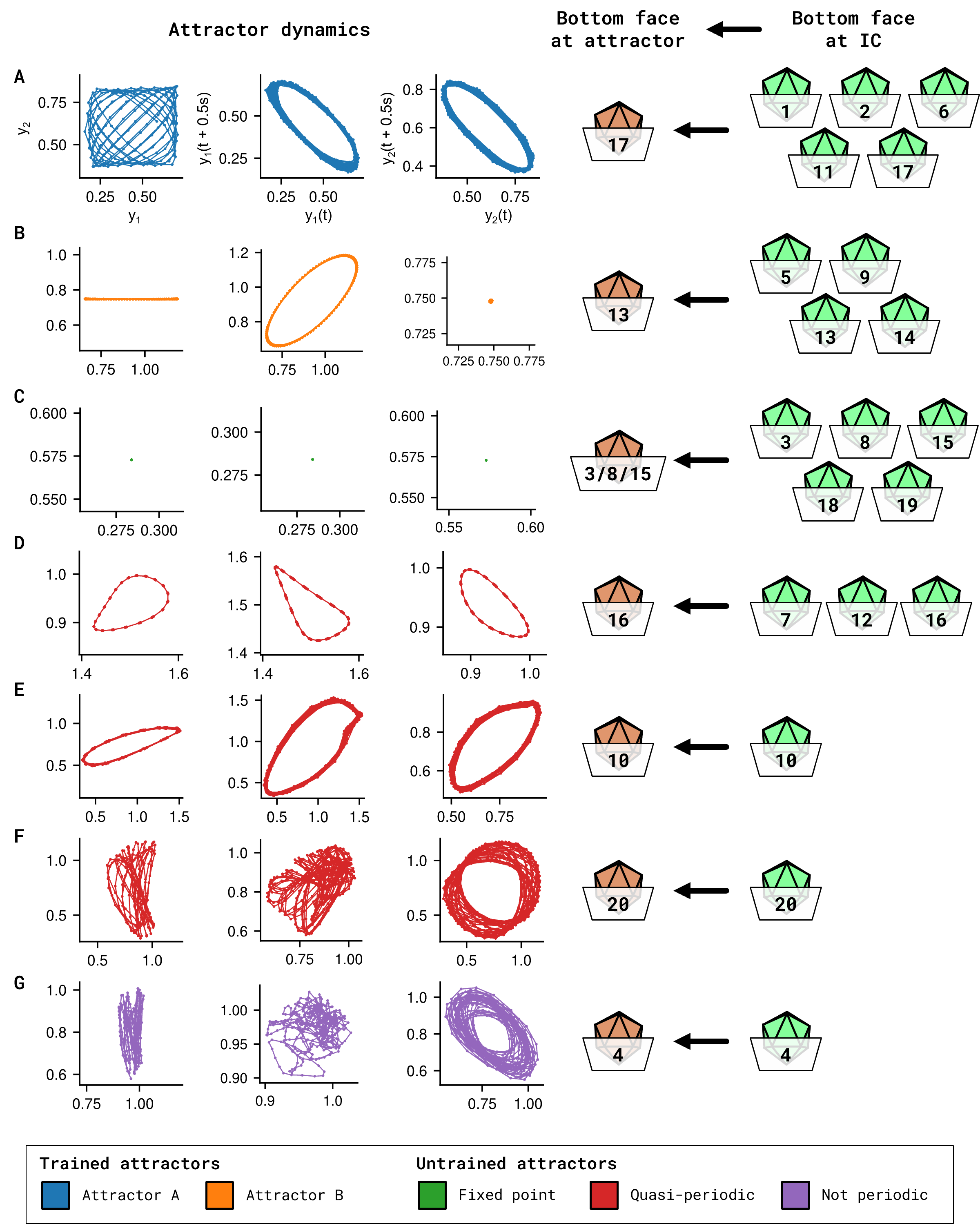}
    \caption{\label{fig:contact} Attractors of the trained system obtained from different initial conditions, corresponding to each bottom face of the robot. The plots are reservoir outputs displayed in the $(y_1, y_2)$ plane, time-delay embedding plots of $y_1$, and time-delay embedding plots of $y_2$. A total of seven attractors is observed, labeled as Attractors A--G. Attractors A and B are trained attractors, whereas Attractors C--G are untrained attractors.}
\end{figure*}

\begin{figure}
    \includegraphics[width=0.9\linewidth]{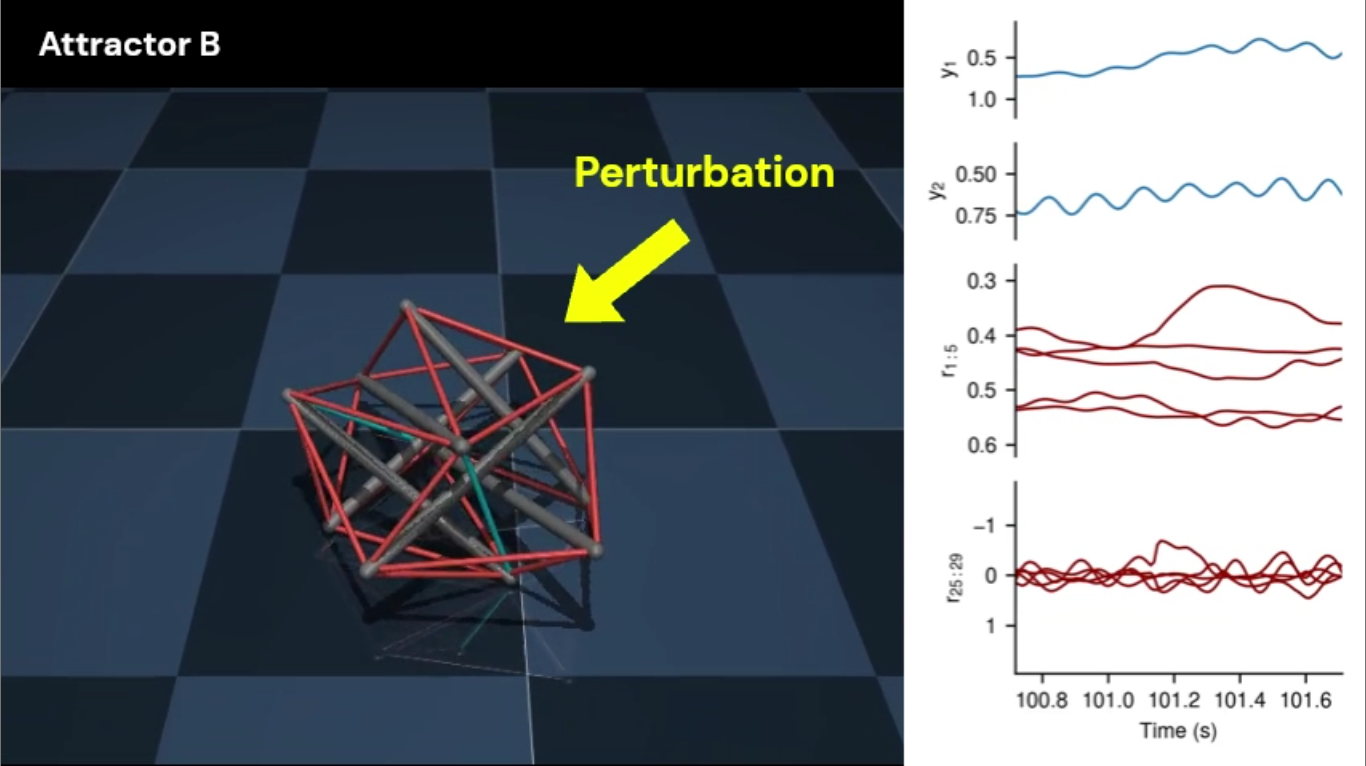}
    \caption{\label{fig:video} Snapshot of the behavior switching demonstration. The robot is switched from Attractor B to Attractor A by forcing a change in its bottom face with an external perturbation (Multimedia available online). The video also includes an archive of behaviors for Attractors A--G.}
\end{figure}

\subsection{Influence of the robot's bottom face}
\label{sec:result_untrained}
In the previous section, two attractors were successfully reconstructed by MF-PRC, corresponding to the ICs of $\bm{s}_\mathrm{A}^\mathrm{I}$ and $\bm{s}_\mathrm{B}^\mathrm{I}$. We call them \emph{Attractor A} and \emph{Attractor B}, respectively. Naturally, the next question to ask is whether the system converges to one of these two attractors from any IC, or if there are other attractors that exist in its state space. To investigate this, we conduct a series of closed-loop simulations from different ICs and observe the attractors that emerge. The same trained readout weights from Sec.~\ref{sec:result_recon} are used in all simulations.

The tensegrity robot used in the current study can be seen as an icosahedron, a polygon with 20 triangular faces. Therefore, the states of the robot can loosely be categorized into 20 regions, depending on which face is in contact with the ground. In this experiment, we define 20 different ICs where the robot sits completely still with each face as the \emph{bottom face}, the face that is in contact with the ground. These ICs are different from the ICs introduced earlier, $\bm{s}_\mathrm{A}^\mathrm{I}$ and $\bm{s}_\mathrm{B}^\mathrm{I}$, which are the states after transient behavior dies out in the open-loop setting. Also, the robot subsequently moves from its motionless initial state because the trained feedback starts to drive the actuators. We label each face of the icosahedron with a number from \{1\} to \{20\}, as shown in Fig.~\ref{fig:icosa}.

Fig.~\ref{fig:contact} shows the complete list of attractors that are observed in this series of closed-loop simulations, along with the bottom faces of the robot that correspond to each attractor. In each case, the output of the physical reservoir (i.e., the feedback control signal) is plotted in the $(y_1, y_2)$ plane and as time-delay embedding plots for $y_1$ and $y_2$. The two-dimensional output dynamics is a good representation of the entire system dynamics because the current system is globally coupled. In addition, the colors of the plots are chosen according to the type of dynamics shown in each attractor. The colors are:
\begin{enumerate}
    \item Blue---If the reconstruction error (considering time shifts) between the generated output and the target $\bm{d}_\mathrm{A}(t)$ is below a specific threshold, we consider that the system has converged to Attractor A.
    \item Orange---If the reconstruction error (considering time shifts) between the generated output and the target $\bm{d}_\mathrm{B}(t)$ is below a specific threshold, we consider that the system has converged to Attractor B.
    \item Green---If the generated output remains constant during the final 2,000 steps of simulation, we consider that the system has converged to a fixed-point attractor.
    \item Red---If the maximum value of the autocorrelation function of the generated output is above a specific threshold, we consider that the system has converged to a periodic or quasi-periodic attractor.
    \item Purple---If none of the above conditions is met, we consider that the system has converged to a non-periodic attractor.
\end{enumerate}
Further details of the categorization method are provided in Appendix~\ref{app:attractor}.

Attractor A is reproduced when the bottom face of the robot is \{17\}. From other ICs of \{1, 2, 6, 11\}, the system produces transient dynamics that cause the robot to roll or tumble until its bottom face becomes \{17\} and it converges to Attractor A. Therefore, this attractor is observed in five out of the 20 ICs. Within this attractor, the robot always shows the same crawling behavior as described previously.

Similarly, Attractor B is reproduced when the bottom face is \{13\}. From other ICs of \{5, 9, 14\}, the system produces transient dynamics that rolls to \{13\} and converges to Attractor B. Within this attractor, the robot shows a small oscillatory movement with no locomotion distance.

From ICs of \{3, 8, 15, 18, 19\}, the system converges to a single fixed-point attractor, represented as \emph{Attractor C} in Fig.~\ref{fig:contact}. Within this attractor, the tensegrity robot relaxes to a specific state and stays completely still. At its relaxed state, the robot is deformed in a way that multiple neighboring faces of \{3, 8, 15\} are in contact with the ground simultaneously.

A particular periodic attractor is observed when the bottom face is \{16\}, represented as \emph{Attractor D} in Fig.~\ref{fig:contact}. From other ICs of \{7, 12\}, the system produces transient dynamics that rolls to \{16\} and converges to Attractor D. Within this attractor, the robot shows a small oscillatory movement with no locomotion distance, similar to that of Attractor B. The subtle difference is that both actuators are actively moving in Attractor D, while only one actuator is moving in Attractor B. Moreover, the amplitude of motion is smaller in Attractor D.

A slightly noisy periodic attractor is observed when the bottom face is \{10\}, represented as \emph{Attractor E} in Fig.~\ref{fig:contact}. Note that despite the noisy time series, there is no stochasticity involved in the current system. Within this attractor, the robot shows a rapid crawling motion that does not efficiently move forward, but instead goes around in circles. Its movement is more intense than that of Attractor A, due to larger amplitudes of $y_1$ and $y_2$.

A quasi-periodic attractor is observed when the bottom face is \{20\}, represented as \emph{Attractor F} in Fig.~\ref{fig:contact}. In terms of robot behavior, Attractor F is similar to Attractor E but is slightly less intense, with the robot showing a rapid crawling movement that goes around in circles.

Finally, a non-periodic attractor is observed when the bottom face is \{4\}, represented as \emph{Attractor G} in Fig.~\ref{fig:contact}. This attractor is the only one that maintains a low value of the autocorrelation function for arbitrary time delays. Because the current system, composed of the physics engine and the trained readout, is fully autonomous and deterministic, Attractor G is likely to be a chaotic attractor. The robot displays an irregular trembling or shaking motion with no locomotion distance.

\begin{figure*}
    \includegraphics[]{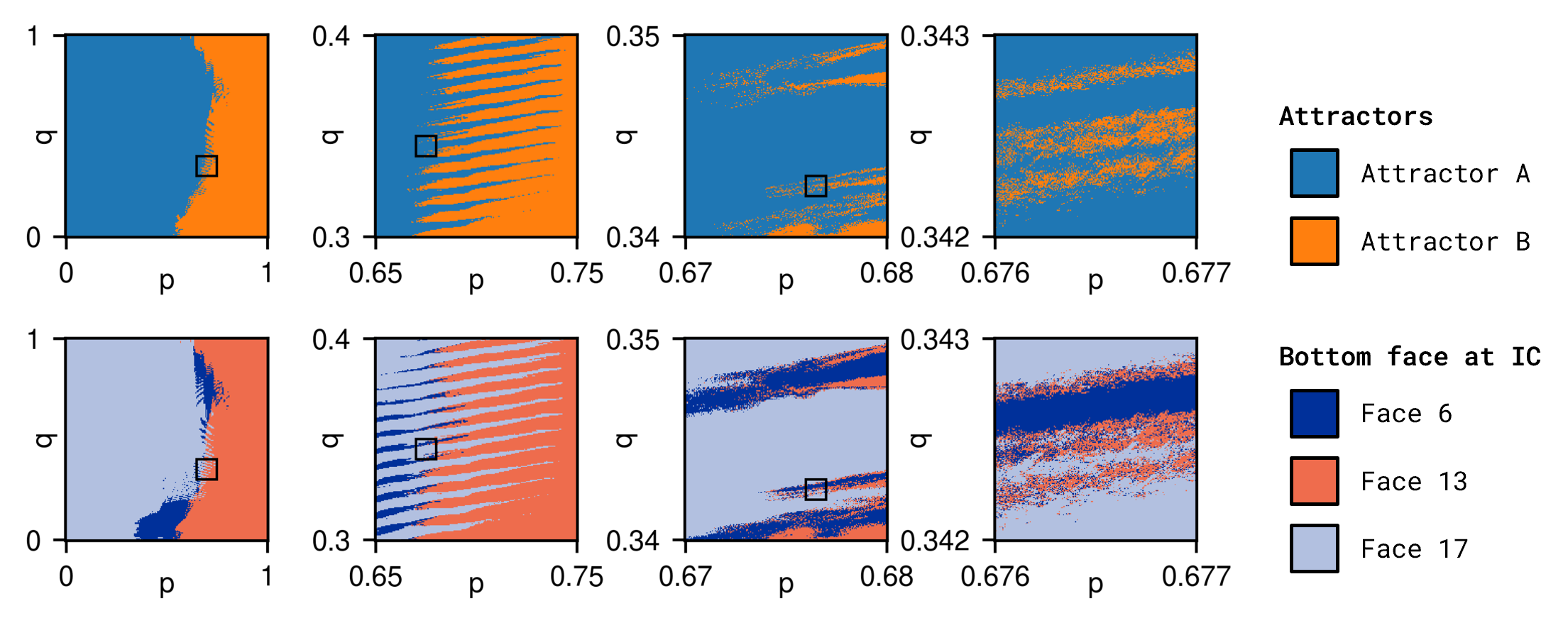}
    \caption{\label{fig:basin} Investigation of attractor basins. The trained system is initialized from various ICs parameterized by $(p, q)$, and the system converges to different attractors depending on the IC. The ICs of the closed-loop simulations are defined as the final states of the open-loop simulations, where the robot is driven by the input signal $\bm{u}_{p,q}(t)$. The upper row shows which attractor the system converges to from each IC. The lower row shows the bottom face of the robot at each IC. From left to right in each row, the plot is zoomed in near the basin boundary. The black box in each plot indicates the parameter region of the next plot.}
\end{figure*}

Among these observed dynamics, Attractors A and B are \emph{trained attractors} because they are explicitly given during the training of MF-PRC as target signals. On the other hand, Attractors C--G are \emph{untrained attractors}, which are not originally intended results but are byproducts of the whole training process. While it is difficult to predict beforehand what types of untrained attractors emerge from the MF-PRC training, they are constrained by multiple factors, including the choice of targets, the learning algorithm, and intrinsic properties of the physical reservoir, that is, the robot body and the environment. For example, the time series of $y_1$ and $y_2$ in all attractors except Attractor C (the fixed point) have timescales similar to the given targets and the robot displays oscillatory motions. The change in frequency of the target signals would also be accompanied by a change in the timescale of untrained attractors.

Importantly, the robot's bottom face is a key factor that determines which attractor the robot converges to. Other investigations (such as the results of Sec.~\ref{sec:result_basin}) suggest that the resulting attractor of the system is more dependent on the bottom face than the initial movement of the robot. Taking advantage of this property, the robot behavior can easily be switched from one to another by changing its bottom face in some way, such as by applying an external perturbation. Such a demonstration is shown in Fig.~\ref{fig:video}, where the robot is switched from Attractor B to Attractor A (Multimedia available online). A plausible explanation is that the state space of the robot is loosely partitioned into regions corresponding to each bottom face, preventing multiple attractors from interfering with each other. The morphological properties of the robot are reflected in the attractor landscape of the trained MF-PRC system and can be exploited to achieve multifunctionality. In the current case, we achieved a coexistence of seven attractors, including two trained attractors and five untrained attractors. We can expect to increase the number of trained attractors and conduct a more intricate design of multistability by increasing the number of target signals and explicitly specifying the bottom face for each target, during the collection of training data.

\subsection{Investigation of attractor basins}
\label{sec:result_basin}
In the previous section, we observed that the robot converges to different attractors from different bottom faces. However, there is also a possibility that the robot converges to different attractors from different ICs with the same bottom face. Therefore, we investigate the attractor basins of the system in more detail by sampling ICs in an alternative way.

There are two points to consider when searching ICs of the current system. Firstly, the dynamical state of the system is high-dimensional and exploring the entire state space becomes expensive. Secondly, the system state must always follow realistic physical constraints; for example, we must be careful not to choose states in which the rigid bars of the tensegrity robot are penetrating the ground or each other. To avoid these problems, we initialize the robot by driving it with an external input signal for a long period of time, and then using its final state as the IC for the closed-loop simulation. The driving input is parameterized by two parameters $(p, q)$, and therefore we can consider the domain of ICs organized in the $(p, q)$ plane. A similar method was proposed in previous studies on MF-RC \cite{flynn2023seeing}.

More specifically, the driving input signal is expressed by the following equations:
\begin{equation}
    \label{eq:basin_input}
    \bm{u}_{p,q}(t) =
    \left[
        \begin{array}{r}
            a_p \sin (\omega_p t + \phi_p) + b_p \\
            a_q \sin (\omega_q t + \phi_q) + b_q
        \end{array}
    \right],
\end{equation}
where:
\begin{equation}
    \label{eq:basin_param}
    \left\{
        \begin{aligned}
            a_p &= (1-p)a_1 + p a_3 \\
            a_q &= (1-q)a_2 + q a_4
        \end{aligned}
    \right.
\end{equation}
and similarly for $\omega_p, \omega_q, \phi_p, \phi_q, b_p,$ and $b_q$. Note that parameters such as $a_1$ are taken from Table~\ref{tab:phenotype}; therefore, the driving input signal $\bm{u}_{p,q}(t)$ can be seen as an equation-level interpolation between the original target signals $\bm{d}_\mathrm{A}(t)$ and $\bm{d}_\mathrm{B}(t)$. The robot is driven by $\bm{u}_{p,q}(t)$ in the open-loop configuration for 20,000 steps, and its final state $\bm{s}^\mathrm{I}$ is used as the IC for the successive closed-loop simulation. The closed-loop simulation is conducted for 80,000 steps to ensure convergence of the system to attractor states. When $p = q = 0$, the input signal is equivalent to target $\bm{d}_\mathrm{A}(t)$ and therefore, $\bm{s}^\mathrm{I} = \bm{s}_\mathrm{A}^\mathrm{I}$. Under this condition, the system converges to Attractor A. Similarly, when $p = q = 1$, the system converges to Attractor B. Fig.~\ref{fig:basin} shows the attractors from other ICs of $\bm{s}^\mathrm{I}$ for parameters $(p, q)$ within the range of $[0, 1]$.

The results show that all ICs from this parameter space converge to either Attractor A or B. The upper-left colormap in Fig.~\ref{fig:basin} indicate that parameter $p$ has a stronger influence on the attractors, forming a basin boundary at roughly $p = 0.65$. However, a closer look into the basin boundary reveals that the boundary is quite complex, with many small-scale structures. The upper-right colormap in Fig.~\ref{fig:basin} shows that in some regions near the basin boundary, the resulting attractor is sensitive to small changes in parameters $p$ and $q$.

However, it cannot be concluded from these results alone that the resulting attractor is sensitive to small changes in the system's ICs. Rather, it is revealed that the distribution of ICs in the $(p, q)$ plane is not continuous. The lower colormaps of Fig.~\ref{fig:basin} show the bottom face of the robot at each IC in the $(p, q)$ plane. The comparison between the upper and lower colormaps indicates that the basin boundary previously discussed is equivalent to the boundary of bottom faces at the IC. ICs with bottom faces of \{6, 17\} converge to Attractor A and ICs with bottom faces of \{13\} converge to Attractor B. These results further support the conclusion that the robot's bottom face is the primary factor that determines the attractor of the closed-loop MF-PRC system.

The small-scale structures observed in the basin boundary are indications that the final state of the \emph{open-loop} system is highly sensitive to parameters of the driving input signal. For clarification, the initial state of the open-loop system is kept constant for all experiments, with the robot staying still on the ground with \{1\} as the bottom face. Therefore, the difference in the final state of the open-loop simulation (that is, the IC of the closed-loop simulation) is solely due to the difference in the input. Previous studies have shown that final state sensitivity\cite{grebogi1983final} is inherent in the act of dice tossing under some certain conditions, contributing to the unpredictability of the dynamic process\cite{nagler2008random,kapitaniak2012three}. Another study has demonstrated that a similar phenomenon occurs in a passive dynamic walker, a mechanical system that mimics bipedal walking\cite{akashi2019unpredictable}. The current results suggest that the locomotion behavior of tensegrity robots with external motor signals can also exhibit a similar type of sensitivity. This sensitivity could be exploited for practical purposes, such as when we want to select the behavior of the robot probabilistically. By selecting input parameters $(p, q)$ within specific regions near the basin boundary, we can emulate a pseudo-stochastic process through a deterministic mechanism, especially because in reality, it is impossible to specify the parameter values or drive the actuators with infinite precision. In Appendix~\ref{app:sensitivity}, we conduct a more thorough analysis of this sensitivity.

\begin{figure*}
    \includegraphics[]{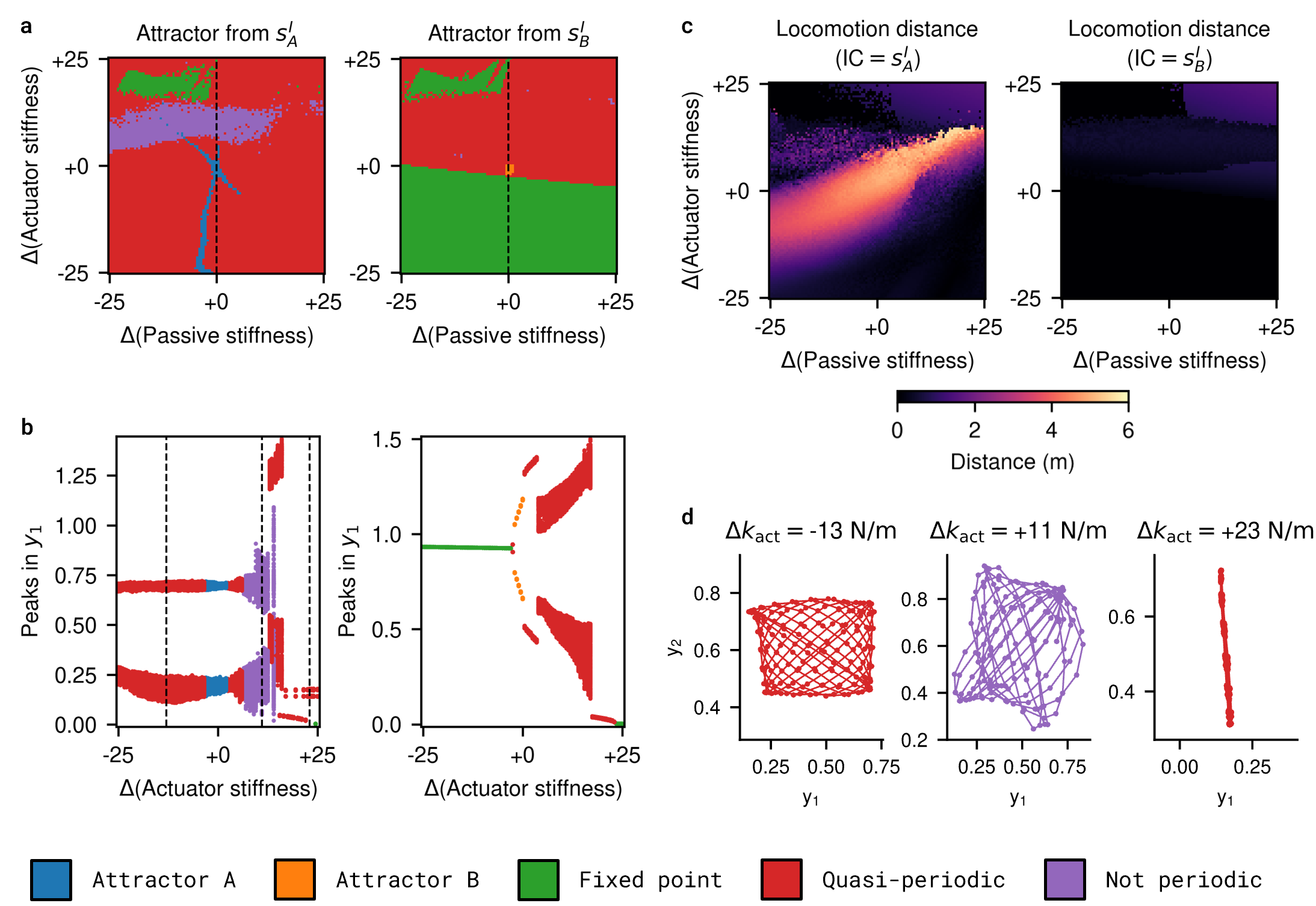}
    \caption{\label{fig:post_learning} Investigation of the system's response to changes in tendon stiffness, after training. (a) A colormap that categorizes the system's attractor from ICs of $\bm{s}_\mathrm{A}^\mathrm{I}$ and $\bm{s}_\mathrm{B}^\mathrm{I}$. The horizontal axis represents the amount of change in the stiffness of passive tendons, and the vertical axis represents the amount of change in the stiffness of actuators. (b) Bifurcation diagram along the slice $k_\mathrm{pas} = 375$~N/m. The vertical axis represents the locally maximum and minimum points of $y_1$ in each attractor. (c) The robot's locomotion distance over 1,000 steps, produced by its attractor dynamics. (d) Examples of untrained attractors sampled from the bifurcation diagram (b). Their positions in the bifurcation diagram are indicated by the dotted lines.}
\end{figure*}

\subsection{System response to body parameter changes}
\label{sec:result_adapt}
As previously discussed, the robot behaviors are attractors of a dynamical system and they are robust to small perturbations in system states. In real-world scenarios, however, robots may receive perturbations not only in terms of system states, but also in terms of system parameters. For example, the change in ground texture can be considered as a change in environment parameters. Similarly, damage, injury, or the growth of some body parts can be seen as a change in body parameters. Therefore, the robustness of robot behaviors should be understood by the system's response to changes in both system states and system parameters.

In this section, we investigate the response of the closed-loop MF-PRC system to changes in some body parameters of the robot. Specifically, we choose as the body parameters, the stiffness of the passive tendons $k_\mathrm{pas}$ and the stiffness of the actuators $k_\mathrm{act}$. The original stiffness values are 375~N/m for both passive tendons and actuators. First, we take the same closed-loop MF-PRC system as the previous experiments; in other words, we take the multifunctional readout trained on targets with phenotype parameters of Table~\ref{tab:phenotype}. Second, the system is initialized with ICs of $\bm{s}_\mathrm{A}^\mathrm{I}$ and $\bm{s}_\mathrm{B}^\mathrm{I}$. Third, we sample a pair of $(k_\mathrm{pas}, k_\mathrm{act})$ within the range from 350~N/m to 400~N/m, change the stiffness of all passive tendons to $k_\mathrm{pas}$, and change the stiffness of both actuators to $k_\mathrm{act}$. Finally, we conduct the closed-loop simulation for 80,000 steps and record the resulting attractor. We repeat this process for each point of $(k_\mathrm{pas}, k_\mathrm{act})$ within the specified range.

Fig.~\ref{fig:post_learning}(a) shows the complete map of attractors in the $(k_\mathrm{pas}, k_\mathrm{act})$ plane. The label of the axes represent the change in stiffness values, compared with the original value of 375~N/m. As expected, the system converges to Attractor A from $\bm{s}_\mathrm{A}^\mathrm{I}$ and to Attractor B from $\bm{s}_\mathrm{B}^\mathrm{I}$ near the middle of the colormap, where the tendons' stiffness is close to the original value.
However, these regions of Attractors A and B are limited to a rather small portion of the parameter space. In other regions, the system converges to untrained attractors of various types. Notably, the system initialized with $\bm{s}_\mathrm{B}^\mathrm{I}$ converges to a fixed-point attractor in most cases when the stiffness of actuators is decreased from the original value.

Fig.~\ref{fig:post_learning}(b) shows the bifurcation diagram along the slice $k_\mathrm{pas} = 375$~N/m, which is represented by a dotted line in Fig.~\ref{fig:post_learning}(a). The vertical axis of this bifurcation diagram represents all of the locally maximum and minimum points of $y_1$ in each attractor. In closed-loop simulations initialized with $\bm{s}_\mathrm{A}^\mathrm{I}$, the system reconstructs Attractor A with high accuracy within the range $372 \leq k_\mathrm{act} \leq 377$~N/m. Although the reconstruction error is high outside this region, the system still displays qualitatively similar dynamics for $k_\mathrm{act} \leq 381$~N/m. Around $k_\mathrm{act} = 381$~N/m, there is a bifurcation taking place from a quasi-periodic attractor to a chaotic attractor. Beyond $k_\mathrm{act} = 388$~N/m, a few different attractors are intertwined in parameter space. In closed-loop simulations initialized with $\bm{s}_\mathrm{B}^\mathrm{I}$, we observe a bifurcation from a stable fixed point to a stable periodic orbit resembling Attractor B near $k_\mathrm{act} = 372.5$~N/m, which suggests that this is a supercritical Hopf bifurcation. There are further bifurcations between different periodic and quasi-periodic attractors at $k_\mathrm{act} = 375.5$~N/m, 379~N/m, and 392.5~N/m. Towards the right end of the bifurcation diagram, we observe the appearance of a stable fixed point for both ICs.

These results indicate that the accurate reconstruction of the target attractors by the closed-loop MF-PRC system is only robust against small changes in tendon stiffness. However, this does not necessarily indicate that robot behaviors are also sensitive to small changes in body parameters. Fig.~\ref{fig:post_learning}(c) shows the locomotion distance measured during the final 1,000 steps of closed-loop simulation, in the same parameter space as Fig.~\ref{fig:post_learning}(a). In simulations initialized with $\bm{s}_\mathrm{A}^\mathrm{I}$, there is no clear correlation between the system's success in reconstructing Attractor A and its locomotion performance. Rather, the locomotion performance of the robot is maintained in a large region of the parameter space. If the underlying motivation behind PRC is to control the locomotion behavior of soft robots, it is more important that the dynamics of the state space are working as intended, rather than the accuracy of the dynamics of the output space. From this view, the reconstruction of target signals is merely a temporary objective that guides the desired behavior of the PRC system. (However, in other tasks of PRC outside the context of locomotion control, this particular view may not be appropriate.)

\begin{figure*}
    \includegraphics[]{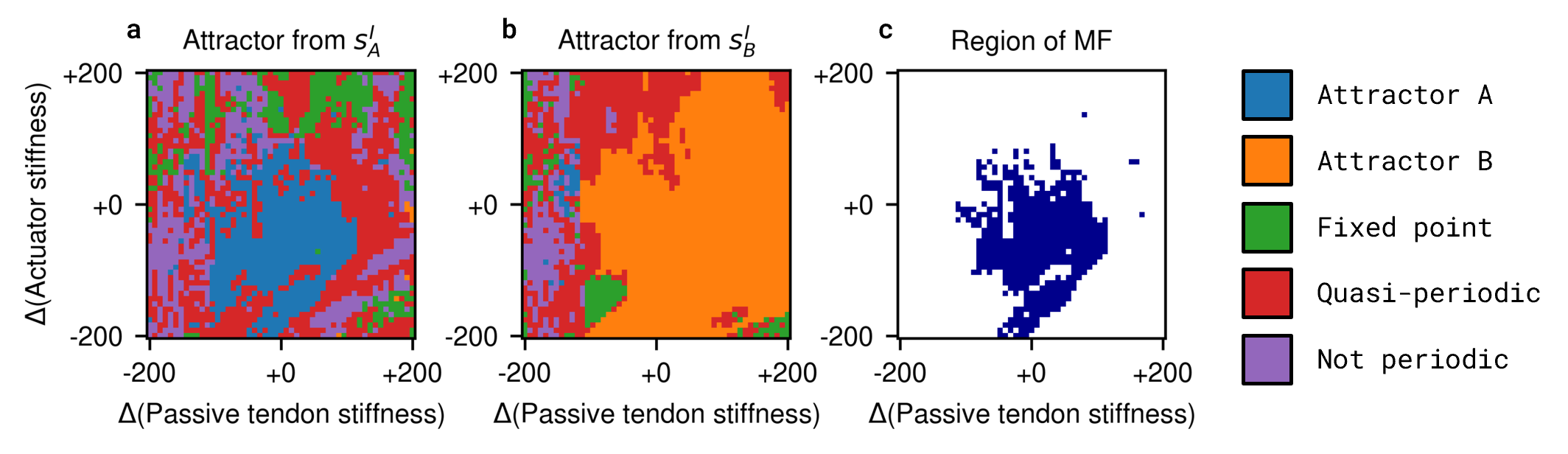}
    \caption{\label{fig:regionmf} The whole MF-PRC training process is repeated for each point of $(k_\mathrm{pas}, k_\mathrm{act})$. Horizontal and vertical axis values represent the difference in stiffness compared to 375~N/m, which is the default condition. (a) Attractor reconstruction results for the IC of $\bm{s}_\mathrm{A}^\mathrm{I}$. The system successfully replicates Attractor A in the blue regions. (b) Attractor reconstruction results for the IC of $\bm{s}_\mathrm{B}^\mathrm{I}$. The system successfully replicates Attractor B in the orange regions. (c) Dark blue areas indicate the region of multifunctionality, where the reconstruction of both Attractors A and B is successful.}
\end{figure*}

With a large change in body parameters, the MF-PRC system clearly goes through bifurcations and converges to untrained attractors. In terms of robot behavior, these untrained attractors are qualitatively different from the trained attractors. As shown in Fig.~\ref{fig:post_learning}(d), they are also different from untrained attractors discussed in Sec.~\ref{sec:result_untrained}. The bifurcations observed here are caused by parameter changes of the PRC system, after the readout is trained and the closed-loop is formed. Therefore, these bifurcations can be called \emph{post-learning bifurcations} \cite{terasaki2024thermodynamic}. We discuss the implications of untrained attractors obtained by post-learning bifurcations further in Sec.~\ref{sec:discussion}.

\subsection{Regions of multifunctionality}
\label{sec:result_bif}
Let us briefly recall that the basic objective of MF-RC and MF-PRC is to achieve the accurate reconstruction of multiple target attractors using a single reservoir system (an ANN or a physical reservoir) and a single set of readout weights. However, the success of training is dependent on the reservoir's dynamical properties and its compatibility with the chosen target attractors. A previous study on MF-RC using echo state networks investigated the dependence of training success on the network's spectral radius, which is an important parameter that determines the dynamics of the network. The parameter regions in which the training is successful are called \emph{regions of multifunctionality} \cite{flynn2021mf}. In this section, we investigate the region of multifunctionality for the tensegrity robot.

We choose again the stiffness of the passive tendons and the actuators as the body parameters to investigate. The parameters $(k_\mathrm{pas}, k_\mathrm{act})$ are set to a specific value and the whole MF-PRC procedure is conducted from the beginning, including the open-loop simulation, the training of the multifunctional readout, and the closed-loop simulation. The crucial difference of the current experiment, compared with the previous experiment in Sec.~\ref{sec:result_adapt}, is that the training is repeated from scratch for each parameter value of $(k_\mathrm{pas}, k_\mathrm{act})$. The targets of training are the same as the previous experiments; refer to Table~\ref{tab:phenotype} for information about the targets. After training, the ICs of the closed-loop simulations are defined as the final states of the open-loop simulations, $\bm{s}_\mathrm{A}^\mathrm{I}$ and $\bm{s}_\mathrm{B}^\mathrm{I}$. Note, however, that the actual values of the ICs are different for each point of $(k_\mathrm{pas}, k_\mathrm{act})$, unlike the previous experiment. We conduct the closed-loop simulations for 80,000 steps and record their resulting attractors. The result of this analysis is shown in Fig.~\ref{fig:regionmf}.

The blue area in Fig.~\ref{fig:regionmf}(a) indicates the region of parameter values in which the reconstruction of Attractor A is possible. The orange area in Fig.~\ref{fig:regionmf}(b) indicates the region of parameter values in which the reconstruction of Attractor B is possible. This area is larger for Attractor B, covering high values of $k_\mathrm{pas}$. The overlapping regions of the blue area in Fig.~\ref{fig:regionmf}(a) and the orange area in Fig.~\ref{fig:regionmf}(b) indicate the conditions where both attractors can successfully be reconstructed and multifunctionality is achieved by the PRC system. This region of multifunctionality is explicitly shown in Fig.~\ref{fig:regionmf}(c). Note that the value range of $(k_\mathrm{pas}, k_\mathrm{act})$ is from 175~N/m to 575~N/m in this experiment, which is much wider than the previous experiment in Sec.~\ref{sec:result_adapt}. If the MF-PRC system is allowed to repeat the training process from scratch, the system can express multifunctionality in a relatively wide range of body parameters.

Outside the region of multifunctionality, the system converges to untrained attractors from one or both of the ICs. The distribution of untrained attractors is rather complex, with stable fixed points, periodic attractors, and strange attractors scattered throughout the parameter space. This suggests that various bifurcation structures exist in the background of MF-PRC training. In the previous experiment, we observed post-learning bifurcations that occur because of parameter changes after the training of the readout. In contrast, the bifurcations observed in the current experiment are caused by parameter changes before training, because the training process is repeated for each point of $(k_\mathrm{pas}, k_\mathrm{act})$. Therefore, the bifurcations seen in Fig.~\ref{fig:regionmf} can be called \emph{pre-learning bifurcations} \cite{terasaki2024thermodynamic}.

\section{Discussions}
\label{sec:discussion}
The results presented here are first implementations of MF-PRC on (simulated) soft robots. The system is capable of learning multiple target motor signals that generate distinct robot behaviors, and these behaviors are obtained as attractors of a dynamical system that are robust to perturbations in state space. In this section, we discuss the implications of this MF-PRC system in the context of embodied AI research.

\subsection{Attractor dynamics in embodied intelligence}
\label{sec:dis_multi}
Constructing the behavior of robots as dynamical attractors provides simple and robust solutions to various control problems. For example, the artificial potential field method \cite{khatib1986real} addresses the path-planning problem of autonomous mobile robots by formulating a vector field with the goal as a stable fixed point. In this method, adaptive behaviors such as obstacle avoidance can be achieved simply by adding some constraints to the equation of the vector field. Another control approach called dynamic movement primitives \cite{schaal2006dynamic,ijspeert2013dynamical,saveriano2023dynamic} introduces a canonical dynamical system with adjustable weights as the basic unit of motion control. These primitives have guaranteed properties of convergence to fixed-point or limit-cycle attractors, and they can be adjusted to meet task-specific requirements. The common theme in these methods is that there is no direct planning of the robot's behavior trajectory. Rather, these methods define the set of constraints that the robot must satisfy, and the actual behavior is self-organized as a result of the system's time evolution.

These approaches align naturally with the view that embodied intelligent behavior is a result of close interactions between the controller, the body, and the environment \cite{pfeifer2007self,pfeifer2006body,pfeifer2014cognition,kuniyoshi2019fusing,kuniyoshi2024embodiment}. In other words, the rules of the controller, the physical constraints of the body, and the dynamics of the environment are all integrated to form a coupled dynamical system \cite{beer1995dynamical}. A great demonstration of this principle is locomotion control using central pattern generators (CPGs). CPGs are neural circuits found in both invertebrate and vertebrate animals that generate coordinated rhythmic patterns while only receiving non-rhythmic inputs \cite{ijspeert2008central}. They are often modeled as systems of coupled oscillators.
Taga et al. demonstrate that bipedal locomotion of a humanoid can be self-organized by a global entrainment between the rhythmic activities of the CPG and the musculoskeletal body \cite{taga1991self}. Ijspeert et al. implemented CPGs in a salamander-like robot and explained the mechanism of automatic transitions between walking and swimming locomotion \cite{ijspeert2007swimming}. Inspired by these works, CPG-based control has been implemented in various tensegrity robots as well \cite{bliss2012central,mirletz2015goal,caluwaerts2013locomotion,caluwaerts2014design,fujita2018environmental}. In these examples, the state trajectory of robot behavior is not directly planned or computed by the controller, but is instead self-organized through interactions with the body and the environment. The limit-cycle dynamics offer advantages such as robustness to perturbations and reduction of control dimensionality.

Put differently, the functionality of control is distributed throughout the controller, the body, and the environment \cite{collins2001three,beal2006passive}. The framework of PRC extends this statement by suggesting that computational capabilities such as nonlinearity and short-term memory, which are typically provided by ANNs in machine learning, can also be distributed to the physical dynamics. Nakajima et al. demonstrate that a physical reservoir composed of soft silicone arms can approximate nonlinear functions that require short-term memory of past inputs \cite{nakajima2014exploiting,nakajima2015information,nakajima2018exploiting}. In the context of the current study, the generation of Lissajous curves cannot be achieved by the linear readout layer alone and is also dependent on the computational capabilities of the tensegrity robot. In practical terms, computation using physical dynamics is considered to be powerful in various robotic applications, such as the design and implementation of fast reflexes, improvement in energy efficiency, and enhancement of sensory capabilities \cite{tanaka2021flapping,yu2023tapered,sakurai2022durable,nakano2023kinesthetic}. There are also various metrics that can be used to measure the computational capability of physical bodies \cite{jaeger2001short,dambre2012information,kubota2021unifying}.

MF-PRC, investigated in this paper, shows that multistability is another appealing property that can be provided by physical reservoirs. In the case of our tensegrity robot, there is a clear relationship between the robot's bottom face and its attractor. Depending on its bottom face, the robot converges to distinct behaviors including consistent forward locomotion, oscillatory motions on the spot, and completely still states (stable fixed points). Therefore, this robot can exhibit multiple functionalities and can easily be switched back and forth between these behaviors. 

In previous studies, multiple-attractor behavior control has been achieved with a combination of bifurcation control and parameter modulation. In the aformentioned study on salamander robots, a one-dimensional input of the CPG acts as a bifurcation parameter that switches the locomotion style between walking and swimming \cite{ijspeert2007swimming}. According to the work by Tani et al., recurrent neural networks with parametric biases can memorize multiple attractors of robot behavior and organize them in the low-dimensional parameter space \cite{tani2004self}. The work by Steingrube et al. uses a chaotic CPG to generate the behavior of a hexapod robot, and a parameter representing the period of orbit is used as a bifurcation parameter to switch between distinct behaviors \cite{steingrube2010self}. While the modulation of bifurcation parameters is an effective method for manipulating the system dynamics, it does not address multistability in the strict sense, where the system converges to different attractors due to changes in the ICs alone.

On the other hand, the focus of our method is to demonstrate that multistability can be designed in physical systems without relying on system parameter changes. Distinct behaviors are self-organized from different ICs, using the same set of readout weights as the controller. At this point, the readout weights can be seen as a simple form of a tunable neural network. The current results indicate that the same neural network can be used to control vastly different robot behaviors, if state trajectories of the body and the environment are separated for each behavior. Instead of training or preparing an independent module for each unit of behavior, it is possible or sufficient to train a versatile neural network that is responsible for all the behaviors. This statement clearly presents advantages in terms of computational efficiency. The more important question probably is what dynamical properties, structures, or constraints of the body and the environment allow this to happen. We propose that MF-PRC is a promising framework that can be used to investigate this question for a variety of physical reservoir setups, with different robot morphologies and environments.

\subsection{Exploiting untrained attractors}
\label{sec:dis_dev}
In attractor reconstruction, the task of RC systems (including also PRC systems, in this section) is to learn the input-output mapping of the evolution function and replicate the target attractor's ergodic properties. In some conditions, the trained system may fail to reconstruct the target attractor, instead converging to an attractor that is unseen during the training phase. These new attractors are called untrained attractors \cite{flynn2021mf,flynn2023seeing} and are byproducts of the whole training process. Although it is difficult to predict the exact nature of untrained attractors, they are constrained by the target choices, the learning algorithm, and properties of the reservoir system. In fact, there are various scenarios in which untrained attractors appear. Below, we provide a brief description of these scenarios.

Firstly, the trained system may globally converge to one or more untrained attractors when the training fails and the system is unable to reconstruct the target attractors. This happens most commonly in closed-loop RC when the acquired input-output mapping is unstable; an error in the output produces an even greater error, causing the system to diverge from the target attractor. In the case of an ESN with a $\tanh$ activation function, its dynamics are bounded by design and must converge eventually to an untrained attractor \cite{flynn2023theory,o2025confabulation}. This is not necessarily the case in physical reservoirs; indeed, in the current study, a small fraction of phenotypes in early stages of the genetic algorithm causes the physics simulator to explode. Realistically, these types of divergence should be suppressed, for example, by clipping the range of actuator signals within a certain range. In such cases, an untrained attractor is expected to emerge due to mechanisms similar to the case of ESNs.

Next, a successfully trained RC system may converge to untrained attractors from ICs that do not exist in the training data. In the current study, the multifunctional readout is trained using data collected from a limited number of simulation runs (one run each for Simulations A and B). Therefore, the training data consists mostly of robot states with bottom faces of \{17\} (from Simulation A) and \{13\} (from Simulation B). As a result, the closed-loop system converges to trained attractors from ICs with these bottom faces but may converge to untrained attractors from other ICs, as described in Sec.~\ref{sec:result_untrained}.

Tracking the bifurcation structure of trained and untrained attractors with respect to system parameters reveals other interesting scenarios. As previously discussed, these bifurcations can be classified into post-learning bifurcations and pre-learning bifurcations \cite{terasaki2024thermodynamic}.

Post-learning bifurcations are caused by changes in system parameters of the closed-loop RC system after the readout is trained. Sec.~\ref{sec:result_adapt} of the current study introduced untrained attractors caused by changes in the tendons' stiffness and which are emergent results of MF-PRC training. On the other hand, numerous studies on RC have shown that post-learning bifurcations can be controlled or anticipated \cite{terasaki2024thermodynamic,kong2021machine,fan2021anticipating,kim2021teaching,patel2021using,fan2022learning,tadokoro2024trans,o2025confabulation}. (Some researchers call these anticipated attractors, \emph{generated attractors}, and distinguish them from untrained attractors \cite{o2025confabulation}.) In a study by Kim et al., an ESN with an additional input channel is trained using data of transient trajectories that converge to fixed points of the Lorenz system. Following a special training procedure, the trained ESN can extrapolate the bifurcation structure of the target Lorenz system and reconstruct the chaotic dynamics of the famous butterfly-shaped attractor \cite{kim2021teaching}, even though it has never seen the chaotic attractor during training. Terasaki and Nakajima demonstrate that a simple ANN model trained on discrete period-three orbits can go through a series of post-learning bifurcations to produce orbits of all periods \cite{terasaki2024thermodynamic}. To summarize, ANN-based RC systems can display surprising extrapolation capabilities through post-learning bifurcation. The application of these methods to PRC systems, including soft robots, is an interesting research direction.

Pre-learning bifurcations are caused by changes in system parameters prior to the training of the readout. In Sec.~\ref{sec:result_bif}, we observed that the MF-PRC system can successfully learn both target attractors in a certain range of body parameters but generates untrained attractors outside this range. In many cases, the trained weights of the readout layer change continuously with respect to changes in the system parameters. Therefore, the untrained attractors obtained through pre-learning bifurcations are not arbitrary attractors but are often dynamically connected to the target attractors. For example, Kabayama et al. point attention to the fact that ESNs commonly display a pre-learning bifurcation from a successful reconstruction of the target attractor to chaos, when the spectral radius of the network is increased. Near the bifurcation point, it is possible to obtain a chaotic attractor that maintains a similar shape to the target attractor \cite{kabayama2024designing}. If this method is applicable to PRC-based control of soft robots, it may be possible to achieve coordinated behavior and chaos simultaneously, allowing these robots to avoid deadlock situations (similar to the hexapod robot controlled by chaotic CPGs \cite{steingrube2010self}).

Untrained attractors are ubiquitous phenomena that occur in all dynamical systems that learn, and yet they have traditionally been dismissed as failures or artifacts of the training process. However, as discussed above, there is increasing evidence that untrained attractors contain structures connected to the targets of training and intrinsic properties of the learning system. In the context of the current study, the untrained attractors observed from different bottom faces of the robot in Sec.~\ref{sec:result_untrained} suggest that the morphological structure of the robot determines the distribution of trained and untrained attractors in the state space. More specifically, it is likely that the robot's state space is loosely partitioned into regions that are associated with different bottom faces, and that this property is important for the coexistence of Attractors A--G. In addition, these attractors operate on a relatively similar timescale that is inherited from the target motor signals. Further analysis of the basin structure and the transient dynamics may reveal additional connections between the trained and untrained attractors.

It is also interesting to investigate and compare attractor landscapes obtained by MF-PRC using different body structures. In this study, the placement of actuators are deliberately chosen to break the symmetry of the robot, based on preliminary observations that such placements allow the generation of more diverse behaviors. By conducting MF-PRC experiments with various actuator placements, we can evaluate the effects of body symmetry in attractor learning. Similarly, counting the number of attractors that can be learned by a larger tensegrity structure with more faces may provide insights related to the body's scale and complexity. According to discussions in embodied cognitive science, body morphologies play a crucial role in shaping the information structure that can be learned through sensorimotor interactions \cite{lungarella2005information,pfeifer2007self,kuniyoshi2019fusing,kuniyoshi2024embodiment}. Based on this idea, a musculoskeletal model of a human fetus has been developed to investigate the highly specific dynamical structures that are induced by human-like body morphologies \cite{kuniyoshi2019fusing,kuniyoshi2024embodiment}. In comparison, the current study on tensegrities enables the analysis of more basic morphological properties such as symmetry and body orientation with respect to the ground (the bottom faces). These vastly different studies form a complementary relationship for the understanding of attractor landscapes formed by learning in embodied systems.

An imperative goal in the field of deep learning is to improve the task generalization capability of robots. Concepts such as transfer learning \cite{zhu2023transfer} and zero-shot learning \cite{jang2022bc} have been proposed to highlight how action policies trained on a certain group of tasks can generalize to perform well in other tasks. When the learning machine, the robot body, and the task environment are viewed together as a coupled dynamical system, the evaluation of the system's performance in a new task setting is analogous to the analysis of untrained attractors. Robots trained with MF-PRC can act as simpler models of sophisticated robotic systems that allow tractable analyses of dynamical structures obtained as a result of the learning process.

Furthermore, developmental robotics \cite{lungarella2003developmental,asada2009cognitive,cangelosi2010integration,taniguchi2023world} emphasizes that intelligent embodied behavior is acquired through autonomous and continuous learning processes based on sensorimotor interactions. In other words, learning that occurs at a particular point in time is always dependent on what has previously been learned. The key idea here is that the previously learned content includes not only successful outcomes of training, but also byproducts and incorrectly learned behaviors. Therefore, untrained attractors are equally important as trained attractors, since they both impose heavy constraints on the system's future learning. From this point of view, untrained attractors acquired at one stage of learning may prove useful in subsequent stages, for the generation and exploration of novel behaviors. MF-PRC in a robotic system is a good framework for investigating this hypothesis.

\subsection{Relation with parameter-aware RC}
\label{sec:dis_parc}
In some recent work on multiple attractor reconstruction using RC, an additional input channel is introduced to provide a unique label for each target attractor. After training, the trained attractors can easily be retrieved by providing the corresponding label as input to the RC system \cite{o2025confabulation,kong2024reservoir,du2024multi,inoue2020designing}. This approach is sometimes called \emph{parameter-aware RC} \cite{kong2021machine}. In relation to the study of memory retrieval in brain and computer systems, parameter-aware RC is associated with location-addressable memory, where each label of the target attractor acts like a memory address that retrieves the desired attractor. In contrast, in the MF-RC and MF-PRC approaches introduced in this paper, a target attractor is retrieved by providing the IC of the system. The IC can be provided, for example, by conducting an open-loop simulation that drives the RC system using samples of the desired attractor. Since the retrieval of an attractor requires partial information acquired from the same attractor, these approaches are associated with content-addressable memory \cite{kong2024reservoir,du2024multi}.

In the current PRC system of a tensegrity robot, the robot behavior can be selected in the same way; i.e., by driving it in open-loop with the target signals before closing the feedback loop. However, this defeats the purpose of MF-PRC as a control approach, which is to replace the open-loop control with closed-loop control using the multifunctional readout. Instead, knowledge of the basin structure of the system (i.e., the relationship between ICs and the resulting attractors) can be used to select the desired attractor by choosing an appropriate IC. In Sec.~\ref{sec:result_untrained}, we observed that there is a fixed relationship between the bottom face of the robot and the resulting attractor, and that an external perturbation that changes the bottom face can effectively switch the robot behavior (Fig.~\ref{fig:video}). In addition, as a topic of future work, it may be possible to design the basin structure of the closed-loop system to some extent by appropriately selecting the target motor signals or the training data. For example, we could design an artificial set of training data that specifies the mapping between the bottom faces and the target attractors.

It is also possible to implement a physical version of parameter-aware RC, as already demonstrated in artificial muscles \cite{akashi2024embedding}. In the case of our tensegrity robot, the most straightforward way of introducing a parameter-aware input channel is to add another actuator that accepts the labels as motor input. However, in PRC systems, there is more freedom in what to consider as inputs to the physical system. The attractor labels could be encoded in body parameters such as the stiffness of tendons, or in environment parameters such as the strength of gravity. In the former case, the trained robot may adaptively switch its behaviors in accordance with the degradation of the tendons. In the latter case, the robot may change its behavior depending on whether it is operating on Earth or on the Moon. Applications of MF-PRC and parameter-aware RC to achieve such kinds of adaptive multifunctionality are interesting directions for future work.

The main focus of the current study is to demonstrate a simple method of designing multistability using PRC, and to conduct dynamical analysis of the resulting system. To facilitate a simple and general analysis, we employ a relatively simple training procedure and do not introduce a parameter-aware input channel. As discussed above, the MF-PRC approach can be extended to improve the controllability or the adaptability of the robot system. Many of the implications discussed in this paper regarding basin structures, untrained attractors, and bifurcations can be applied to such extended systems as well.

\section{Conclusion}
\label{sec:conclusion}
In this paper, we presented experimental results of MF-PRC on a simulated tensegrity robot, demonstrating that multistability can be designed in physical systems with an appropriately trained linear readout layer. The main focus of this study is the locomotion control of soft robots that fully exploits the morphological properties of the body. Notably, the same trained weights of the readout layer can be re-used to control multiple robot behaviors which are coexisting attractors, as long as their dynamical trajectories are separated in the high-dimensional state space. In some situations, this finding may eliminate the need of modular control architectures, such as introducing a separate readout for each unit of behavior, and reduce the computational load required to generate complex movements.

Furthermore, investigations of untrained attractors reveal interesting behavior outside the training data. The results of Sec.~\ref{sec:result_untrained}, Sec.~\ref{sec:result_adapt}, and Sec.~\ref{sec:result_bif} illustrate the different scenarios in which untrained attractors can be found, including pre-learning and post-learning bifurcations that are caused by system parameter changes before and after training. This classification of bifurcation scenarios is useful for organizing recent findings in RC that highlights the functionality of reservoir dynamics outside the training data \cite{akashi2024embedding,kabayama2024designing,terasaki2024thermodynamic,kong2021machine,fan2021anticipating,kim2021teaching,patel2021using,fan2022learning,tadokoro2024trans,rohm2021model}. Furthermore, untrained attractors are not arbitrary artifacts of the training process, but are always constrained by dynamical properties (or in other words, the embodiment \cite{pfeifer2007self,kuniyoshi2024embodiment}) of the learning system. In Sec.~\ref{sec:dis_dev}, we discussed the importance of this idea in the context of developmental robotics.

Finally, it should be noted that the procedure of MF-PRC is not limited to soft robots, but can be applied to various physical reservoirs, such as photonic, spintronic, and quantum systems. Indeed, a prior work investigates MF-PRC on a photonic reservoir system \cite{zhang2025mf}. Studying MF-PRC on different physical reservoirs is a promising direction for future research, not only because it enhances the performance of these computational systems, but also because it provides further insight into how multifunctionality is related to the "embodiment" of the learning medium.

\begin{acknowledgments}
    This work is supported in part by the Next Generation AI Research Center of the University of Tokyo and the Chair for Frontier AI Research and Education, School of Information Science and Technology, the University of Tokyo.
    R. T. is supported by JSPS KAKENHI Grant Number 23KJ0708.
    K. N. is supported by JSPS KAKENHI Grant Numbers 21KK0182 and 23K18472 and by JST CREST Grant Number JPMJCR2014.
\end{acknowledgments}

\section*{Author Declarations}
\subsection*{Conflict of Interest}
The authors have no conflicts to disclose.

\subsection*{Author Contributions}
\textbf{Ryo Terajima:} Conceptualization (equal); Formal analysis (equal); Funding acquisition (equal); Investigation (lead); Methodology (equal); Software (lead); Writing - original draft (lead); Writing - review \& editing (equal).
\textbf{Katsuma Inoue:} Formal analysis (equal); Investigation (supporting); Methodology (equal); Software (supporting); Writing - review \& editing (supporting).
\textbf{Kohei Nakajima:} Conceptualization (equal); Funding acquisition (equal); Methodology (equal); Supervision (lead); Writing - review \& editing (equal).
\textbf{Yasuo Kuniyoshi:} Conceptualization (equal); Funding acquisition (equal); Supervision (supporting); Writing - review \& editing (equal).

\section*{Data Availability Statement}
The data that support the findings of this study are available from the corresponding author upon reasonable request.

\appendix
\section{Simulation details}
\label{app:physics}
The overall appearance of the tensegrity robot used in the current study is shown in Fig.~\ref{fig:system}(a). The cylindrical rigid bars are implemented as the \emph{capsule} type object in the MuJoCo physics engine. The passive tendons and actuators are implemented as damped springs with no mass, as represented in Eq.~(\ref{eq:tendon}). They are directly attached to the surface of the rigid bars.

In MuJoCo, the state of the dynamical system $\bm{s}(t)$ is represented by the position, orientation, and velocity of all rigid bodies. The position of each body has three degrees of freedom in the x, y, and z directions. The orientation of each body, represented by a unit quaternion, has four degrees of freedom. The velocity of each body has six degrees of freedom, which includes the linear (x, y, z) and angular (pitch, roll, yaw) velocities. Therefore, each rigid body has a total of 13 degrees of freedom. Since there are six rigid bars in the current robot, the total number of degrees of freedom in the simulation state is 78. 

Although the current simulation takes into account certain aspects of realistic robot implementation, the primary goal of the current study is to demonstrate the feasibility of MF-PRC as a proof of concept. Therefore, some simulation settings are idealized. For example, the tendons used in this study can exert force in both the compressive and tensile directions, which may require a special mechanism to implement in a real robot. In addition, the restlengths of the actuators can be adjusted instantaneously, according to motor signals, in the current simulation setting. We assume no specific material for the rigid bars, and therefore, the MuJoCo default density of 1000~kg/m$^3$ is used. Table~\ref{tab:simulation} summarizes the physical parameter values used in the current study, which are required to fully replicate the experiment results. For different parameter settings, the detailed structure of the attractor landscape is expected to change. However, the main conclusions of this study---that multifunctionality can be achieved and that interesting untrained attractors and bifurcations can be obtained---remain unaffected.

\section{Evolutionary optimization details}
\label{app:evolution}
As explained in Sec.~\ref{sec:evolution}, the pair of target signals used in the training of MF-PRC is optimized using a multi-objective genetic algorithm, more specifically the NSGA-II algorithm \cite{deb2002fast}. This algorithm is implemented using the DEAP library \cite{deap2012}. Each phenotype is represented by 16 parameters of the target Lissajous curves, which take continuous values. The phenotype parameter values are sampled from uniform distributions at the initialization of the first generation and during the mutation process. The value range of the amplitudes $a_i$ is from 0.01 to 0.3, the value range of the angular velocities $\omega_i$ is from $2\pi$ to $40\pi$ (corresponding to frequencies from 1~Hz to 20~Hz), the value range of the phases $\phi_i$ is from 0 to $2\pi$, and the value range of the offsets (centers of oscillation) $b_i$ is from 0.4 to 1.0. The genetic algorithm is run for 100 generations, with a population size of 64. At each generation, new offsprings are created by crossover with a probability of 0.7 and by mutation with a probability of 0.3. The parent and offspring populations are combined, and the best 64 phenotypes are selected for the next generation. We set the size of the offspring population to be 64, so the selection rate is 0.5.

\begin{table}
    \caption{\label{tab:simulation} Physical simulation parameters.}
    \begin{ruledtabular}
    \begin{tabular}{cc}
        Parameter & Value \\
        \hline
        Length of rigid bars & 0.90~m \\
        Radius of rigid bars & 0.025~m \\
        Density of rigid bars & 1000~kg/m$^3$ \\
        Restlength of passive tendons & 0.25~m \\
        Actuator lengths at equilibrium, $L$ & 0.84~m \\
        Stiffness of passive tendons, $k_\mathrm{pas}$ & 375~N/m \\
        Stiffness of actuators, $k_\mathrm{act}$ & 375~N/m \\
        Damping of passive tendons and actuators, $c$ & 7.5~kg/s \\
        Coefficient of tangential friction & 1.0 \\
        Coefficient of torsional friction & 0.005 \\
        Coefficient of rolling friction & 0.0001 \\ 
        Timestep of the physics engine & 1~ms \\
    \end{tabular}
    \end{ruledtabular}
\end{table}

\begin{figure}
    \includegraphics[]{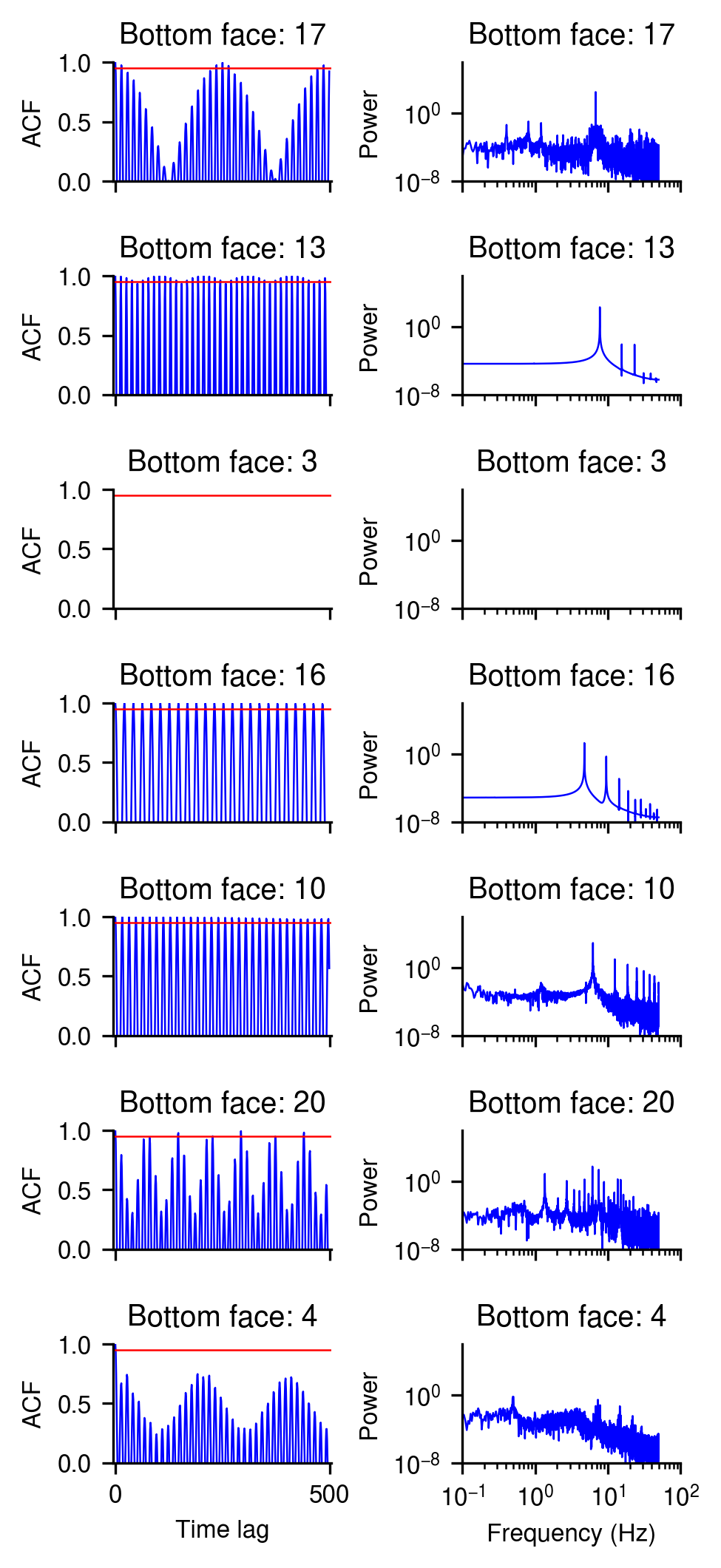}
    \caption{\label{fig:acf} The autocorrelation function of $\bm{y}(t)$ and the power spectrum of $y_1(t)$ for Attractors A--G, as introduced in Fig.~\ref{fig:contact}.}
\end{figure}

\begin{figure*}
    \includegraphics[]{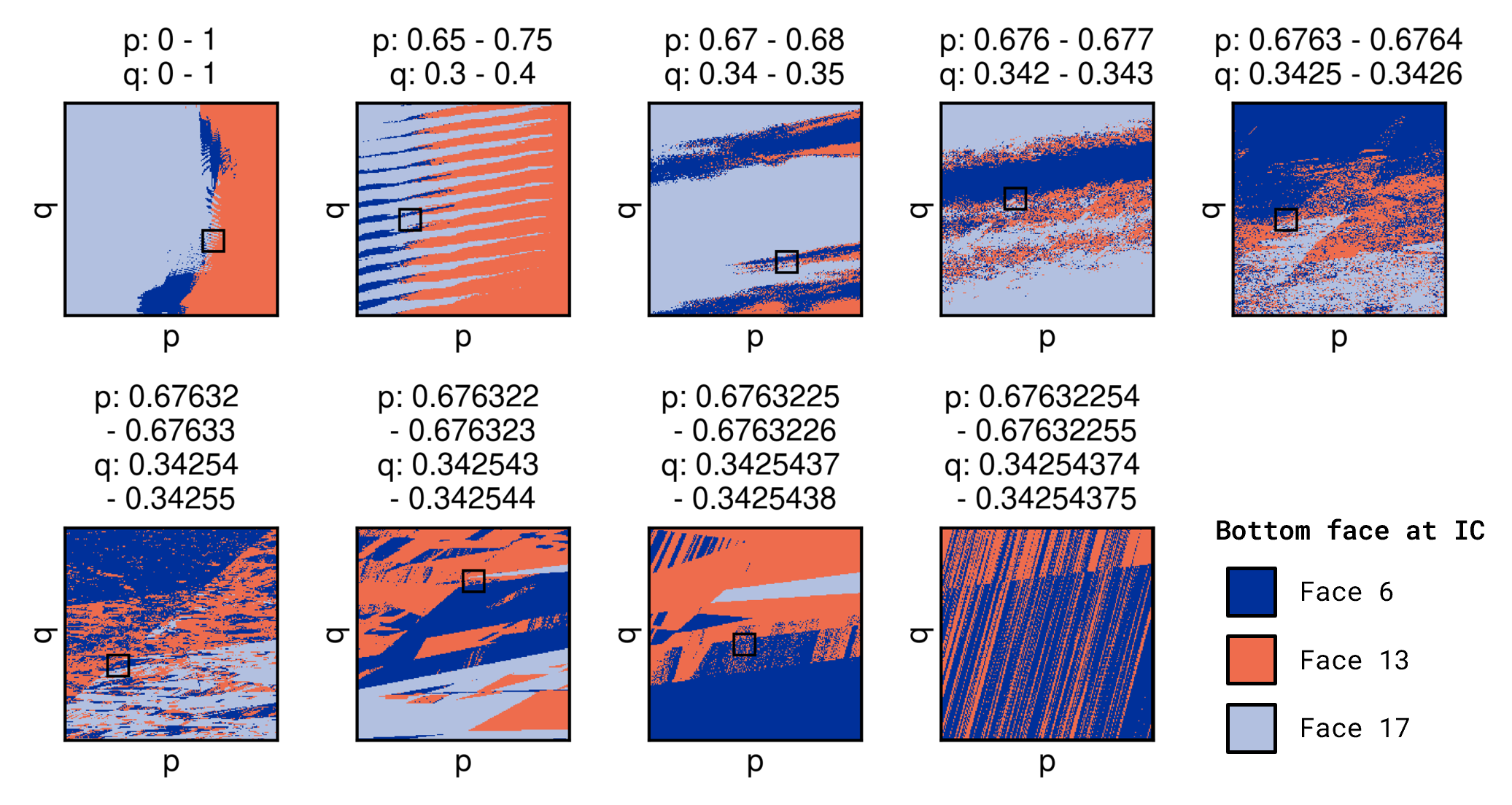}
    \caption{\label{fig:sensitivity} The robot's bottom face at various ICs parameterized by $(p, q)$. From top-left to bottom-right, the investigated region of ICs is zoomed in. The black box in each plot indicates the parameter region of the next plot. The first four plots are identical to what was presented in Fig.~\ref{fig:basin}.}
\end{figure*}

\begin{figure*}
    \includegraphics[]{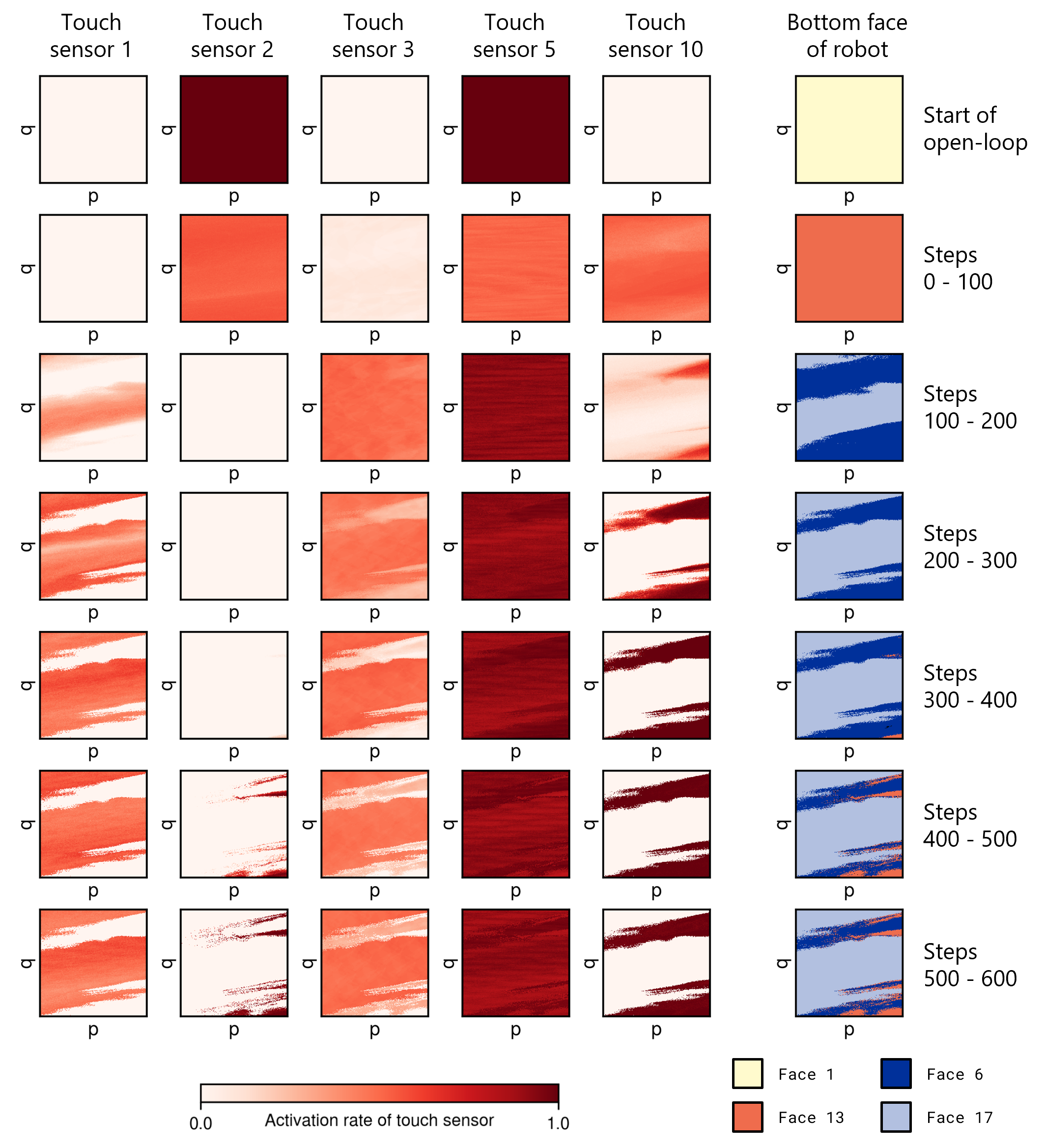}
    \caption{\label{fig:touchrate} The change over time in the robot's touch sensor measurements and its bottom face, during open-loop simulations driven by periodic motor signals parameterized by $(p,q)$. Each row of plots represents a specific time window of the open-loop simulations. The parameter region of interest is taken from the third plot of Fig.~\ref{fig:sensitivity}, which is $0.67 \leq p \leq 0.68$ and $0.34 \leq q \leq 0.35$.}
\end{figure*}

\section{Attractor categorization details}
\label{app:attractor}
In this section, we describe in detail how obtained attractors of the closed-loop MF-PRC system are categorized into different groups. As summarized in Sec.~\ref{sec:result_untrained}, the attractors are cateogized in to five groups: Attractor A (blue), Attractor B (orange), fixed-point attractors (green), periodic and quasi-periodic attractors (red), and other attractors (purple). The categtorization process is conducted in the following order. For convenience, in the following explanations, time $t$ is represented in discrete time, where a single unit of time corresponds to the reservoir timestep, $\tau = 10$~ms.

We first check whether the dynamics can be classified as either Attractor A or Attractor B. This is based on the normalized root mean squared error (NRMSE) between the output time series $\bm{y}(t) = [y_1(t), \, y_2(t)]^\top$ and the target time series $\bm{d}(t) = [d_1(t), \, d_2(t)]^\top$. For each element $i = 1, 2$, the NRMSE with time shifts $\Delta t$ is calculated as:
\begin{align}
    E_i(\Delta t) &=
    \frac{
        \sqrt{\frac{1}{T'}\sum_{t=1}^{T'} (y_i(t + \Delta t) - d_i(t))^2}
    }{
        |\mathrm{max}(d_i(t)) - \mathrm{min}(d_i(t))|
    }, \\
    E(\Delta t) &= \frac{1}{2} \left( E_1(\Delta t) + E_2(\Delta t) \right),
\end{align}
where $T'$ is the length of the time series used in this analysis. If the minimum value of $E(\Delta t)$ for various $\Delta t$ is smaller than a threshold value of $\delta_E$, the attractor is classified as either Attractor A or Attractor B. For this analysis, we used $T' = 6,000$~steps and $\delta_E = 0.30$. We have checked that this choice of $T'$ is sufficiently long, by observing that the classification results do not change drastically with longer $T'$. If the output satisfies the condition $\min_{\Delta t} (E(\Delta t)) < \delta_E$ for both Attractors A and B, the dynamics is classified as the attractor with the smaller error, but this does not generally occur in the current study (because Attractors A and B are different enough).

Next, we identify fixed-point attractors. For this, we obtain the maximum difference of output values within the final $T'$ steps of the simulation:
\begin{align}
    E_i &= \max_t y_i(t) - \min_t y_i(t), \\
    E &= \max_i E_i.
\end{align}
If $E$ is smaller than the threshold $\delta_E$, the attractor is classified as a fixed-point. For this analysis, we used $T' = 2,000$~steps and $\delta_E = 0.01$.

Finally, we consider whether the dynamics can be considered to be either periodic or quasi-periodic. For this, we calculate the autocorrelation function (ACF) of the output time series:
\begin{align}
    \mathrm{ACF}_i(\Delta t) &=
    \frac{
        \sum_{t=1}^{T - \Delta t} (y_i(t) - \bar{y}_i)(y_i(t + \Delta t) - \bar{y}_i)
    }{
        (T - \Delta t)(y_i(t) - \bar{y}_i)^2
    }, \\
    \mathrm{ACF}(\Delta t) &= \frac{1}{2} \left( ACF_1(\Delta t) + ACF_2(\Delta t) \right),
\end{align}
where $T'$ is the length of the time series used in this analysis and $\bar{y}_i$ is the mean of $y_i(t)$. If  the dynamics in question is periodic, the ACF will show peaks with values close to 1 at regular intervals corresponding to its period. On the other hand, if the dynamics is not periodic, the ACF tends to decay with increasing lag. Therefore, we classify the dynamics as periodic or quasi-periodic if the value of ACF is greater than a threshold $\delta_\mathrm{ACF}$ at a certain value of $\Delta t > 50$ steps. For this analysis, we used $T' = 10,000$~steps and $\delta_\mathrm{ACF} = 0.95$. While this is a rather heuristic method of identifying periodic attractors, it allows us to grasp the overall structure of the attractor landscape in our experiments. For reference, Fig.~\ref{fig:acf} displays the ACF and the power spectrum of the output time series for Attractors A--G, introduced in Fig.~\ref{fig:contact}. The threshold value $\delta_\mathrm{ACF}$ is shown as a red line in the figure. Following the above identification method, Attractor G corresponding to the bottom face of \{4\} is the only attractor that is classified as not periodic. This classification is consistent with the lack of distinct peaks in its power spectrum.

\section{Detailed investigation on the sensitivity of the basin structure}
\label{app:sensitivity}
In Sec.~\ref{sec:result_basin}, we investigated the resulting attractor and bottom face of the system from a group of ICs, obtained by driving the system with periodic motor signals. The results of this analysis indicate that the robot's bottom face at the IC is the primary factor that determines the resulting attractor. In addition, in some regions of the parameter space, the bottom face and the resulting attractor are both highly sensitive to small changes in the parameters of the input signal, $(p, q)$. In this section, we provide some supplementary results to support the validity of these conclusions.

In Fig.~\ref{fig:sensitivity}, the investigation of Fig.~\ref{fig:basin} is continued further to focus on narrower regions of the parameter space $(p, q)$. Interestingly, different patterns and features appear depending on the scale of the investigated region. Nevertheless, a smaller structure appears at all scales of the investigation, confirming that the results are in fact sensitive to small changes in the parameters.

The open-loop simulations of this experiment are all initialized with the same system state, and then it is continued for 20,000 steps using the periodic motor signals represented by Eqs.~(\ref{eq:basin_input}) and (\ref{eq:basin_param}). In Figs.~\ref{fig:basin} and \ref{fig:sensitivity}, the bottom face of the robot after 20,000 steps is shown. However, the results of Fig.~\ref{fig:touchrate} reveal that the bottom face mostly converges after only around 600 steps of simulation. Looking at the figure from top to bottom, the distribution of bottom faces in the parameter space, which is uniform at the beginning of the simulation, gradually diverges over time until it reaches a fixed distribution after around 600 steps. This fact is confirmed also by observing the videos of the robot simulations; the behavior and bottom face of the robot becomes stable after around the same time period.

For clarification, in the current study, the bottom face of the robot is identified by measurements of 12 touch sensors that are attached at each end of a rigid bar comprising the robot. Each sensor returns a binary value at every time step. For a given time window, the top three sensors that are most frequently activated are selected, and the triangular face corresponding to these three sensors is identified as the bottom face of the robot. Face \{1\} is represented by sensors 2, 5, and 8; face \{6\} is represented by sensors 3, 5, and 10; face \{13\} is represented by sensors 2, 5, and 10; face \{17\} is represented by sensors 1, 3, and 5. The correspondence between the sensor measurements and the bottom faces can be checked by comparing the left and right plots of Fig.~\ref{fig:touchrate}.

To summarize, the experiment results regarding the basin structure of the system is supported by raw data of the touch sensors and visual inspection of the simulation video. This suggests that the small-scale (and seemingly noisy) features observed in Figs.~\ref{fig:basin} and \ref{fig:sensitivity} are results of physical interactions of the robot, and not numerical artifacts of the simulation. At the same time, the results presented here are unavoidably dependent on assumptions that is inherent in the physics engine MuJoCo, such as its soft contact model and its constraint solver \cite{todorov2012mujoco}. The discussion of whether these assumptions of the physics engine are realistic is beyond the scope of the current study.

\bibliography{bib/main}

\end{document}